\newif\ifeusipstyle
\newif\ifdohybrid
\eusipstylefalse
\dohybridfalse

\ifeusipstyle
  \documentclass[a4paper]{article}
  \usepackage{spconf,amsmath,graphicx,cite}
\else
  \documentclass[conference]{IEEEtran} 

  \usepackage[dvips]{graphicx}
\fi

\usepackage{url}
\usepackage{amsfonts}

\usepackage{floatrow}
\newfloatcommand{capbtabbox}{table}[][\FBwidth]
\usepackage{blindtext}

\title{Trainable Compound Activation Functions for Machine Learning}

\ifeusipstyle
   \name{Paul M Baggenstoss}
   \address{Fraunhofer FKIE, Fraunhoferstr 20,
   \\ 53343 Wachtberg, Germany}
\else
   \author{\IEEEauthorblockN{Paul M. Baggenstoss}
   \IEEEauthorblockA{Fraunhofer FKIE,
   Fraunhoferstrasse 20\\
   53343 Wachtberg, Germany\\
   Email: p.m.baggenstoss@ieee.org}
   }
\fi

\begin{document}
\newcommand{\defined}{\stackrel{\mbox{\tiny$\Delta$}}{=}}
\newtheorem{example}{Example}
\newtheorem{conclusion}{Conclusion}
\newtheorem{assumption}{Assumption}
\newtheorem{definition}{Definition}
\newtheorem{problem}{Problem}
\newcommand{\erf}{{\rm erf}}

\newcommand{\sst}{\scriptstyle }
\newcommand{\xparen}{\mbox{\small$(\bfx)$}}
\newcommand{\hojz}{H_{0j}\mbox{\small$(\bfz)$}}
\newcommand{\Hozj}{H_{0,j}\mbox{\small$(\bfz_j)$}}
\newcommand{\smallmath}[1]{{\scriptstyle #1}}
\newcommand{\Hoz}[1]{H_0\mbox{\small$(#1)$}}
\newcommand{\Hozp}[1]{H_0^\prime\mbox{\small$(#1)$}}
\newcommand{\Hozpp}[1]{H_0^{\prime\prime}\mbox{\small$(#1)$}}
\newcommand{\hoz}{\Hoz{\bfz}}
\newcommand{\hooz}{\Hozp{\bfz}}
\newcommand{\hoooz}{\Hozpp{\bfz}}
\newcommand{\smJ}{{\scriptscriptstyle \! J}}
\newcommand{\smK}{{\scriptscriptstyle \! K}}

\newcommand{\erfc}{{\rm erfc}}
\newcommand{\bitem}{\begin{itemize}}
\newcommand{\dsum}{{ \displaystyle \sum}}
\newcommand{\eitem}{\end{itemize}}
\newcommand{\benum}{\begin{enumerate}}
\newcommand{\eenum}{\end{enumerate}}
\newcommand{\bdm}{\begin{displaymath}}
\newcommand{\bfzro}{{\underline{\bf 0}}}
\newcommand{\bfone}{{\underline{\bf 1}}}
\newcommand{\edm}{\end{displaymath}}
\newcommand{\beq}{\begin{equation}}
\newcommand{\bea}{\begin{eqnarray}}
\newcommand{\eea}{\end{eqnarray}}
\newcommand{\cali}{ {\cal \bf I}}
\newcommand{\caln}{ {\cal \bf N}}
\newcommand{\barray}{\begin{displaymath} \begin{array}{rcl}}
\newcommand{\earray}{\end{array}\end{displaymath}}
\newcommand{\eeq}{\end{equation}}
\newcommand{\btheta}{\mbox{\boldmath $\theta$}}
\newcommand{\bTheta}{\mbox{\boldmath $\Theta$}}
\newcommand{\blam}{\mbox{\boldmath $\Lambda$}}
\newcommand{\beps}{\mbox{\boldmath $\epsilon$}}
\newcommand{\bdelta}{\mbox{\boldmath $\delta$}}
\newcommand{\bgamma}{\mbox{\boldmath $\gamma$}}
\newcommand{\balpha}{\mbox{\boldmath $\alpha$}}
\newcommand{\bbeta}{\mbox{\boldmath $\beta$}}
\newcommand{\balphascript}{\mbox{\boldmath ${\scriptstyle \alpha}$}}
\newcommand{\bbetascript}{\mbox{\boldmath ${\scriptstyle \beta}$}}
\newcommand{\bLambda}{\mbox{\boldmath $\Lambda$}}
\newcommand{\bDelta}{\mbox{\boldmath $\Delta$}}
\newcommand{\bomega}{\mbox{\boldmath $\omega$}}
\newcommand{\bOmega}{\mbox{\boldmath $\Omega$}}
\newcommand{\blambda}{\mbox{\boldmath $\lambda$}}
\newcommand{\bphi}{\mbox{\boldmath $\phi$}}
\newcommand{\bpi}{\mbox{\boldmath $\pi$}}
\newcommand{\bnu}{\mbox{\boldmath $\nu$}}
\newcommand{\brho}{\mbox{\boldmath $\rho$}}
\newcommand{\bmu}{\mbox{\boldmath $\mu$}}
\newcommand{\sigi}{\mbox{\boldmath $\Sigma$}_i}
\newcommand{\bfu}{{\bf u}}
\newcommand{\bfx}{{\bf x}}
\newcommand{\bfb}{{\bf b}}
\newcommand{\bfk}{{\bf k}}
\newcommand{\bfc}{{\bf c}}
\newcommand{\bfv}{{\bf v}}
\newcommand{\bfn}{{\bf n}}
\newcommand{\bfK}{{\bf K}}
\newcommand{\bfh}{{\bf h}}
\newcommand{\bff}{{\bf f}}
\newcommand{\bfg}{{\bf g}}
\newcommand{\bfe}{{\bf e}}
\newcommand{\bfr}{{\bf r}}
\newcommand{\bfw}{{\bf w}}
\newcommand{\calX}{{\cal X}}
\newcommand{\calZ}{{\cal Z}}
\newcommand{\bb}{{\bf b}}
\newcommand{\bfy}{{\bf y}}
\newcommand{\bfz}{{\bf z}}
\newcommand{\bfs}{{\bf s}}
\newcommand{\bfa}{{\bf a}}
\newcommand{\bfA}{{\bf A}}
\newcommand{\bfB}{{\bf B}}
\newcommand{\bfV}{{\bf V}}
\newcommand{\bfZ}{{\bf Z}}
\newcommand{\bfH}{{\bf H}}
\newcommand{\bfX}{{\bf X}}
\newcommand{\bfR}{{\bf R}}
\newcommand{\bfF}{{\bf F}}
\newcommand{\bfS}{{\bf S}}
\newcommand{\bfC}{{\bf C}}
\newcommand{\bfI}{{\bf I}}
\newcommand{\bfO}{{\bf O}}
\newcommand{\bfU}{{\bf U}}
\newcommand{\bfD}{{\bf D}}
\newcommand{\bfY}{{\bf Y}}
\newcommand{\bSig}{{\bf \Sigma}}
\newcommand{\test}{\stackrel{<}{>}}
\newcommand{\zmk}{{\bf Z}_{m,k}}
\newcommand{\zlk}{{\bf Z}_{l,k}}
\newcommand{\zm}{{\bf Z}_{m}}
\newcommand{\ssq}{\sigma^{2}}
\newcommand{\dint}{{\displaystyle \int}}
\newcommand{\ds}{\displaystyle }
\newtheorem{theorem}{Theorem}
\newcommand{\postscript}[2]{ \begin{center}
    \includegraphics*[width=3.5in,height=#1]{#2.eps}
    \end{center} }

\newtheorem{identity}{Identity}
\newtheorem{hypothesis}{Hypothesis}
\newcommand{\mathtiny}[1]{\mbox{\tiny$#1$}}

\maketitle

\begin{abstract}
Activation functions (AF) are necessary components of
neural networks that allow approximation of functions,
but AFs in current use are usually simple monotonically increasing functions.
In this paper, we propose trainable compound AF (TCA)
composed of a sum of shifted and scaled simple AFs.
TCAs increase the effectiveness of networks
with fewer parameters compared to added layers. TCAs have a special
interpretation in generative networks because
they effectively estimate the marginal distributions of each dimension of the data
using a mixture distribution, reducing modality and making linear dimension reduction more effective.
When used in restricted Boltzmann machines (RBMs), they result in a novel
type of RBM with mixture-based stochastic units.
Improved performance is demonstrated in experiments using RBMs, deep belief networks (DBN),
projected belief networks (PBN), and variational auto-encoders (VAE).
\end{abstract}

\section{Introduction}

\subsection{Background and Motivation}
Activation functions (AF) are neccessary components of
neural networks that allow approximation of most types of functions
(universal approximation theory).
Activation functions in current use consist of simple fixed functions
such as sigmoid, softplus, ReLu \cite{FengActFn2019,Ravanbakhsh,WangRelu,jin2016deep}. 
There is motivation to find more complex AFs for machine learning,
such as parametric Relu,
to improve the ability of neural networks to approximate
complex functions or probability distributions \cite{he2015delving}. 
Putting more complexity in activation functions
can increase the function approximation
capability of a network, similar to adding
network layers, but with far fewer parameters.

\subsection{Theoretical Justification}
\begin{figure}[h!]
  \begin{center}
    \includegraphics[width=2.3in,height=2.2in]{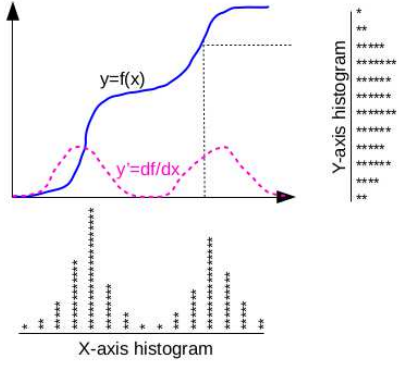}
  \caption{Illustration of an activation function
removing modality in data when the derivative approximates
the histogram.}
\vspace{-.2in}
  \label{modal}
  \end{center}
\end{figure}

Most approaches to selecting AFs focus on the end result, i.e.
performance of the network \cite{FengActFn2019}. 
It may be more enlightening to ask what does 
the AFs say about the input data.  Any monotonically increasing
function can be seen as an estimator
of the input distribution \cite{BagKayInfo2022}.
This view that AFs are PDF estimators can be best described mathematically
with the change of variables theorem.  Let the AF be written $y=f(x)$ and let us assume that 
$y$ is a random variable with distribution $p_y(y)$. Then, 
the distribution of $x$ is given by
\beq
p_x(x)=\left| \frac{\partial y}{\partial x} \right| \;  p_y(f(x)) = \left| f^\prime(x) \right| \;  p_y(f(x)).
\label{pdx}
\eeq
If $y$ has the uniform distribution on $[0,\; 1]$, then 
\beq
p_x(x)= \left| f^\prime(x) \right|.
\label{pdx2}
\eeq
The activation function $f(x)$ can be used as a probability density function
(PDF) estimator if it is adjusted (trained) until $y$ has a uniform output distribution,
so that (\ref{pdx2}) holds.  Training is accomplished by maximum likelihood (ML) 
estimation using
\beq
\max_\theta \frac{1}{K} \left\{  \sum_{i=1}^K \log f^\prime(x_k;\theta)  \right\},
\label{mleq}
\eeq
where $k$ indexes over a set of training samples $x_k$,
and we have removed the absolute value operator because we assume $f(x;\theta)$ is monotonically increasing, so $f^\prime(x;\theta)>0$.
In accordance with (\ref{pdx2}), the trained AF will have increasing slope in regions where the input data $x$ is concentrated,
with the net result being that the output has a uniform distribution.  
This concept is illustrated in Figure \ref{modal}.

A similar argument can be made for a Gaussian output distribution, where
$p_y(y)=\frac{1}{\sqrt{2\pi}}e^{-y^2/2}.$ Then, 
\beq
p_x(x)= \left| f^\prime(x) \right| \frac{1}{\sqrt{2\pi}}e^{-f(x)^2/2}.
\label{pdx3}
\eeq
Training $f(x)$ will then result in $p_y(y)$ approaching the Gaussian distribution.
Some confusion may arise because we are discussing two different distributions of $y$, the true distribution
based on knowing $p_x(x)$, obtained by inverting (\ref{pdx}) given by 
$p_y(y)= \frac{p_x(x)}{\left| f^\prime(x) \right|},$
and the assumed distribution.  The purpose of training $f(x)$ is to make the
true distribution of $y$ approach the assumed distribution.
In general, the slope of $f(x)$ will tend to 
increase where the histogram of $x$ has peaks, serving to remove modalities in the data
as illustrated in Figure \ref{modal}, as the activation function approximates 
the cumulative distribution of the input data.

The view that $f(x)$ is a PDF estimator, and the fact that data often has clusters
leads us to the idea of creating AF's with multi-modal derivatives.
One of the simplest and earliest types of AFs is the sigmoid function,
whose derivative approximates a Gaussian distribution. Therefore,
the sum of shifted sigmoid functions approximates a Gaussian mixture, which
is a popular approach to PDF estimation \cite{Redner,McLachlan}.  Different base activation functions lead to other
mixture distributions.  This view leads us to the idea for the trained compound
activation functions (TCA).

\subsection{Contributions and Goals of Paper}
In this paper, we propose TCA, a trainable activation function with
complex, but monotonic response.  We argue that using a TCA in a neural
network, is a more efficient way to increase the effectiveness of a network
than adding layers.  Furthermore, in generative networks, the TCA has an interpretation 
as a mixture distribution and can remove modality in the data.
When the TCA is used in a restricted Boltzmann machine (RBM),
it creates a novel type of RBM based on stochastic units that are
mixtures. We show significant improvement of TCA-based RBMs, deep
belief network (DBN)  and projected belief networks (PBNs) in experiments.

\section{Trained Compound Activation Function (TCA)}
Consider the compound activation function $f(x)$ given by
\beq
f(x) = \frac{1}{M} \sum_{j=1}^M  \; f_0\left(e^{a_j} x+b_j\right),
\label{tcadef}
\eeq
where $f_0(x)$ is the {\it base} activation function, and  ${\bf a}=\{a_j\}$ and ${\bf b}=\{b_j\}$ 
are scale and bias parameters.  The exponential function $ e^{a_j}$ is used to insure positivity
of the scale factor. Note that if $f_0(x)$ is a monotonically increasing function
(which we always assume), then the TCA is monotonically increasing and equal to the base activation 
function if ${\bf a}={\bf 0}$ and ${\bf b}={\bf 0}$.

For a dimension-$N$ input data vector $\bfx$, the TCA operates element-wise
on $\bfx$  as follows:
\beq
\bfy = f(\bfx), \;\;\; y_i = \frac{1}{M} \sum_{j=1}^M  \; f_0\left(e^{a(i,j)} x_i+b(i,j)\right), \;1\leq i \leq N,
\label{tcadefv}
\eeq
where ${\bf A}=\{a(i,j)\}$ and ${\bf B}=\{b(i,j)\}$ are $N\times M$ scale and bias parameters.  
A TCA can be implemented with an additional dense layer that expands the dimension
to $N\cdot M$ neurons, followed by a linear layer that averages over each group of $M$ neurons,
compressing back to dimension $N$. But, not only does a TCA use a factor of $N$ fewer parameters,
but it has an interpretation as a mixture
distribution when used in generative models, 
and results in a novel type of RBM, as we now show.

\section{TCA for Deep Belief Networks (DBN)}
A deep belief network is a layered network proposed by
Hinton \cite{HintonDeep06} based on restricted Boltzmann machines (RBMs).

\subsection{RBMs}
The RBM is a widely-used generative stochastic artificial neural network
that can learn a probability distribution over its set of inputs
\cite{Goodfellow2016}.  The RBM is based on an elegant stochastic model,
the Gibbs distribution, and is the central idea in a deep belief
network (DBN) made popular by Hinton \cite{HintonDeep06}.
A cascaded series of layer-wise-trained RBMs can be used to initialize deep neural
networks. This method, in fact played a key role in the birth of deep
learning because they provided a means to pre-train deep networks that
suffered from vanishing gradients.  

\subsection{Review of RBMs}
\begin{figure}[h!]
  \begin{center}
    \includegraphics[width=2.7in]{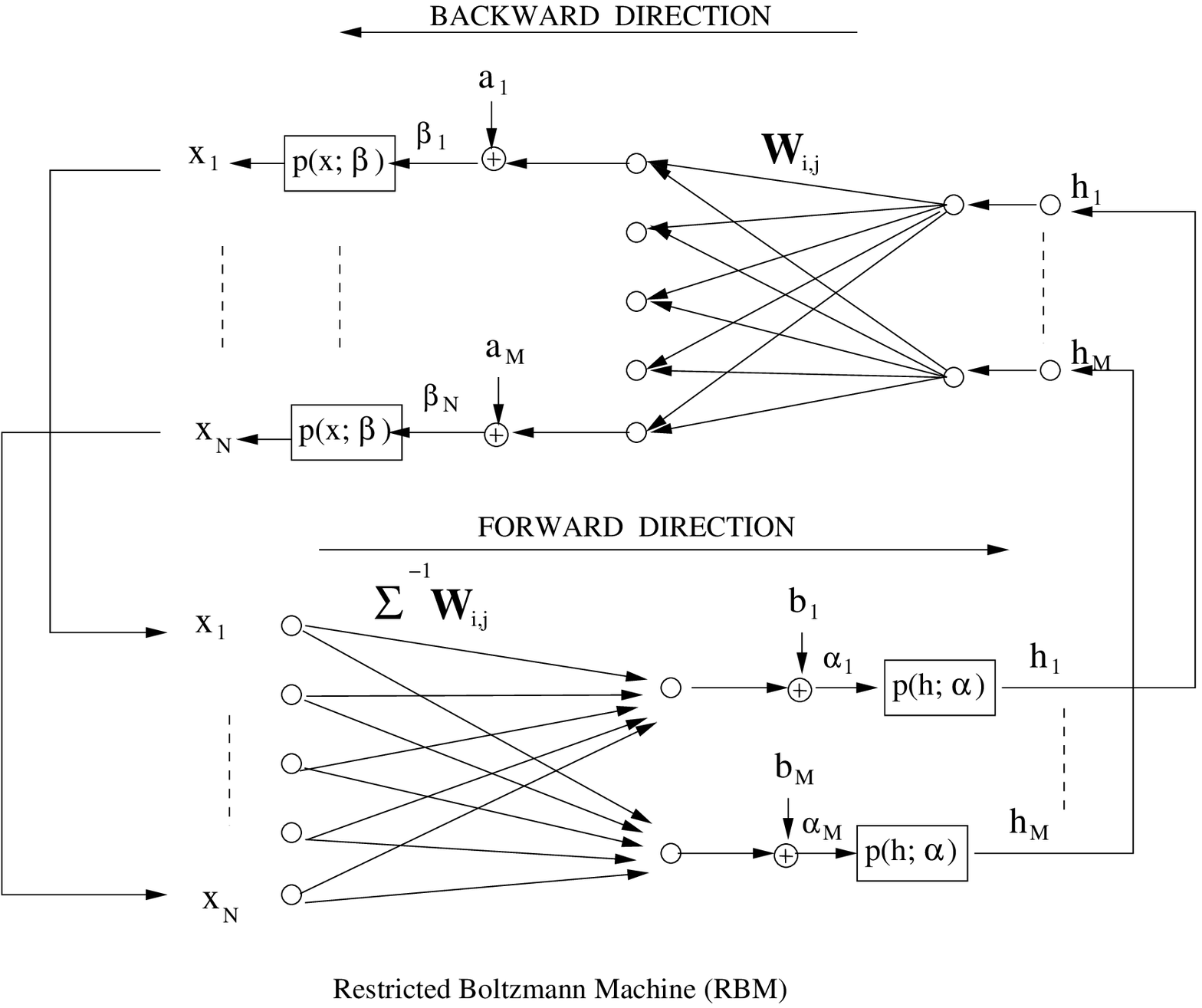}
  \caption{Illustration of an RBM.}
  \label{rbm}
  \end{center}
 \vspace{-.1in}
\end{figure}

The RBM estimates a joint distribution between an input (visible) data vector $\bfx\in\mathbb{R}^N$, and
a set of hidden variables  $\bfh\in\mathbb{R}^M$.
The RBM consists of a pair of stochastic perceptrons, arranged back-to-back,
and is illustrated in Figure \ref{rbm}.
In a sampling procedure called ``Gibbs sampling",
data is created by alternately sampling $\bfx$ and $\bfh$
using the conditional distributions $p_h(\bfh|\bfx)$ and
$p_x(\bfx|\bfh)$.  To sample $\bfh$ from the distribution
$p_h(\bfh|\bfx)$, we first multiply $\bfx$ by the
transpose of the $N \times M$ weight matrix ${\bf W}$, and add a bias vector:
$\balpha = {\bf W}^\prime \bfx + \bfb.$
The variable $\balpha$ is then applied to a generating distribution
(GD) to create the stochastic variable $\bfh$ as $h_i \sim p(h;\alpha_i),$  $1\leq i \leq M$.
Note that conditioned on $\bfx$, $\bfh$ is a set of independent random
variables (RV).  To sample $\bfx$ from the distribution
$p_x(\bfx|\bfh)$, we use the analog of the forward sampling process:
$\bbeta = {\bf W} \bfh + \bfa.$ The variable $\bbeta$ is then applied to a generating distribution
$x_j \sim p(x;\beta_j),$  $1\leq j \leq N$.  Conditioned on $\bfh$,
$\bfx$ is a set of independent random variables (RV).
After many alternating sampling operations, the joint distribution
between $\bfx$ and $\bfh$ converges to the Gibbs distribution
$p(\bfx,\bfh) = \frac{e^{-E(\bfx,\bfh)}}{K},$
where the normalizing factor $K$ is generally unknown.
Training an RBM is done using contrastive divergence, which
is described in detail for exponential-class GDs in \cite{WellingHinton04}.

\subsection{Activation functions and RBMs}
Once an RBM is trained, it can be used as a layer
of a neural network to extract information-bearing features
$\bfh$. This is done by replacing stochastic sampling with deterministic
sampling by replacing the stochastic generating distributions
$p(x;\beta)$, $p(h;\alpha)$ with activation
functions that equal the expected value (mean) of the
generating distributions, $f(x;\alpha)=\mathbb{E}(x;\alpha)$.
Consider the Bernoulli distribution whose AF is the sigmoid function,
the truncated exponential distribution (TED) whose AF is the 
TED distribution \cite{BagEusipcoRBM}, the truncated Gaussian distribution (TG) whose
AF is the TG activation \cite{Bag2021ITG}, and the Gaussan distribution
which has the linear AF $f(x)=x$.

\subsection{RBMs based on TCA}
If a simple activation function corresponds to the expected value
of the GD, then what distribution corresponds to a TCA?
It is previously known that any monotonically-increasing function can be seen as a sum of shifted 
stochastic generating distributions \cite{HintonRelu2010,Ravanbakhsh}.
But, we must look more carefully at this because it is not as simple as
adding random variables.  When adding random variables, the probability densities combine by
convolution, not additively. To combine them properly, we need
a mixture distribution.
Let $p_0(x;\alpha)$ be a univariate generating distribution depending on parameter
$\alpha$, and let this distribution have mean $f_0(\alpha)=\mathbb{E}(x;\alpha)$, so $f_0(\alpha)$
is the AF corresponding to probability distribution $p(x;\alpha)$. 
Let $\Phi_0(x;\alpha)$ be the cumulative distribution function (CDF) of $p_0(x;\alpha)$ , i.e.
$$\Phi_0(x;\alpha)=\int_{-\infty}^x \; p_0(x;\alpha) \; {\rm d} x.$$
Now, consider a mixture distribution given by
\beq
p(x;\alpha)  = \sum_{j=1}^M \;  \frac{1}{M} \;  p_0\left(e^{a_j} \alpha+b_j\right).
\label{tcadefp}
\eeq
To draw a sample from  mixture distribution (\ref{tcadefp}), we first draw a discrete random
variable $j$ uniformly in $[1, \; M]$, then  draw $x$ from
distribution $e^{-a_j} p_0\left(e^{a_j} \alpha+b_j\right)$.
Mixture distribution (\ref{tcadefp}) has CDF
\beq
\Phi(x;\alpha)  = \frac{1}{M} \sum_{j=1}^M \; e^{-a_j} \Phi_0\left(e^{a_j} \alpha+b_j\right).
\label{tcadefP}
\eeq
It is easily seen by taking the derivative, 
that distribution corresponding to the CDF (\ref{tcadefP}) is (\ref{tcadefp}).
And, since expected value is a linear operation,
the mean of distribution (\ref{tcadefp}) is the TCA (\ref{tcadef}).
Note that RBMs are implicitly an infinite mixture distributions
over the hidden variables \cite{Roux2008}, but using 
using discrete mixture $\phi(x;\alpha)$ for a generating distribution
creates an entirely novel type of RBM.

\begin{figure}[h!]
  \begin{center}
    \includegraphics[width=3.0in]{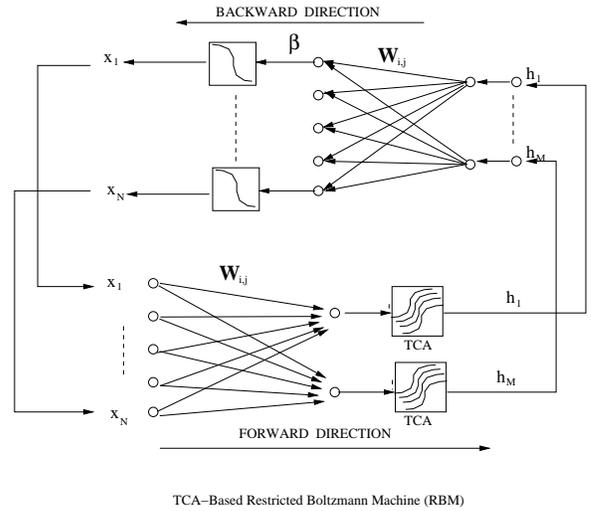}
  \caption{Illustration of an RBM based with TCA units in the
forward path. The need for a separate bias in the forward path is eliminated due
to the existance of trainable bias (shifts).}
  \label{rbm_nnl}
  \end{center}
\end{figure}

Different base AFs (i.e. different base stochastic units) can be used for the input
and output, producing a wide range of different types of RBMs \cite{Bag2021ITG}.
Figure \ref{rbm_nnl} illustrates an RBM constructed using a TCA unit in the forward path.
The activation functions and TCAs in the figure can be either stochastic
(random sampling from the corresponding GD) or deterministic
if the activation functions are used.
In the forward path, a weight matrix ${\bf W}$ multiplies the input data vector ${\bf x}$
in order to produce a linear feature vector, which is then passed
through the TCA to produce the hidden variables vector ${\bf h}$.
In the backward path, ${\bf h}$ is multiplied by the transposed weight matrix ${\bf W}^\prime$
and passed through an activation function to produce the re-sampled input vector ${\bf x}$.
In our approach, we use a TCA only in the forward path,
with a normal AF in the backward path.

The mathematical approach to train the parameters of
RBMs using the contrastive divergence (CD) algorithm 
is well documented \cite{WellingHinton04} and can be 
extended in order to obtain the updates equations to train the 
parameters of the TCAs.  This is facilitated using the symbolic differentiation
available using software frameworks such as THEANO \cite{Theano2010}.


\section{Stacked RBM and DBNs}
To create a ``stacked RBM", an RBM is trained on the input data, and then the
forward path is used to create hidden variables, which are then used as
input data for the next layer.  
The deep belief network (DBN) \cite{HintonDeep06} consists of a series
of stacked RBMs, plus a special ``top layer" RBM.  The one-hot encoded class labels are injected at the input
of the top layer (concatenated with the hidden variables from the last stacked RBM).
Then, the Gibbs distribution of the top layer learns the joint distribution
of the class labels with the hidden variables out of the last stacked RBM.
The cleverness of Hinton's invention lies in the fact that although
the scale factor of the the Gibbs distribution is not known, 
it is not needed to compare the likelihood function from the
competing class hypotheses.
Computing the Gibbs distribution for a given class
assumption has been called the ``free energy" \cite{HintonDeep06,Bag2021ITG},
so we will call this a free energy classifier.
Computing the free energy classifier requires solving for terms
of the marginalized Gibbs distribution \cite{Bag2021ITG},
and these in turn require the CDF, which we have given in (\ref{tcadefP}).
We therefore have all the tools to create a DBN using TCA-based
stochastic units.

%

\section{Experiments: TCA-based RBM and DBN}
\subsection{Data}
\label{ddesc}
For the these experiments,  we took a subset of the  MNIST handwritten data corpus,
just three characters ``3", ``8", and ``9". 
The data consists of sample images of $28\times 28$, or a data dimension of 784.
We used 500 training samples from each character.
Since MNIST pixel data is coarsely quantized in the range [0,1],
a dither was applied to the pixel values\footnote{For pixel values
above 0.5, a small exponential-distributed random value was subtracted,
but for pixel values below 0.5, a similar  random value was added.}.  

\subsection{Network}
The network was a 1-layer stacked RBM of 32 neurons, followed by a top-level (classifier) RBM
of 256 units.  TCA's with 3 components were used in the
forward path.  For the base activation and stochastic unit, 
truncated exponential distribution (TED), which is the continuous version of the 
Bernoulli distribution/sigmoid function \cite{BagEusipcoRBM,Bag2021ITG}, is used.

\subsection{First Layer}
In the first experiment, we trained just the first layer RBM and measured
input data reconstruction error after one Gibbs sampling cycle.
We consider both mean-square error and conditional likelihood function (LF)
which is $\log p(\bfx|\bbeta)$, where $\bbeta$ is the input to the
activation functions in the reconstruction path (see  Figure \ref{rbm_nnl}).

\begin{figure}
\begin{floatrow}
\ffigbox{%
  \includegraphics[width=1.5in]{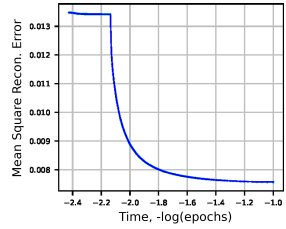}
}{%
     \caption{First layer mean square reconstruction error (MSE) as a function of training epoch
    with log-time in X-axis. After convergence at X-axis location -2.15, the TCAs were allowed to change.}
  \label{L1profile}
}
\capbtabbox{%
     \begin{tabular}{|l|l|l|}
    \hline
    AF & MSE & LF\\
     \hline
    TED   & .0135 & -7.0 \\
     \hline
    TCA-0 & .0134 & -7.0 \\
     \hline
    TCA  & .0029 & -2.59\\
     \hline
    \end{tabular}
}{%
     \caption{MSE and conditional LF for first layer only. TCA-0: initial (but not updated) TCA.}
     \label{tab1aa}
}
\end{floatrow}
\end{figure}

We trained in three phases, (a) first with no TCA (using
just the base AF), then (b) with TCA but with TCA update
disabled, then finally (c) with TCA enabled.
At initialization, the TCAs have a transfer function very similar to the base
non-linearity, so with TCA update disabled, we should expect the same performance as for the base AF. 
Training was done using contrastive divergence \cite{WellingHinton04,HintonDeep06}.
For the first layer, we used deterministic Gibbs sampling
(using AF instead of stochastic units).
When switching from phase (b) to (c), we plotted the
MSE as a function of epochs. In Figure \ref{L1profile},
the plot begins where phase (b) has reached convergence,
then at X axis -2.25, the TCA training is enabled and
a drastic change is seen.
In Table \ref{tab1aa}, we listed the final MSE and LF for the
three phases. Nearly a factor of 2 reduction in MSE is seen. 
The improved reconstruction of TCA can be seen on the bottom row.



\begin{figure}[h!]
  \begin{center}
    \includegraphics[width=3.0in]{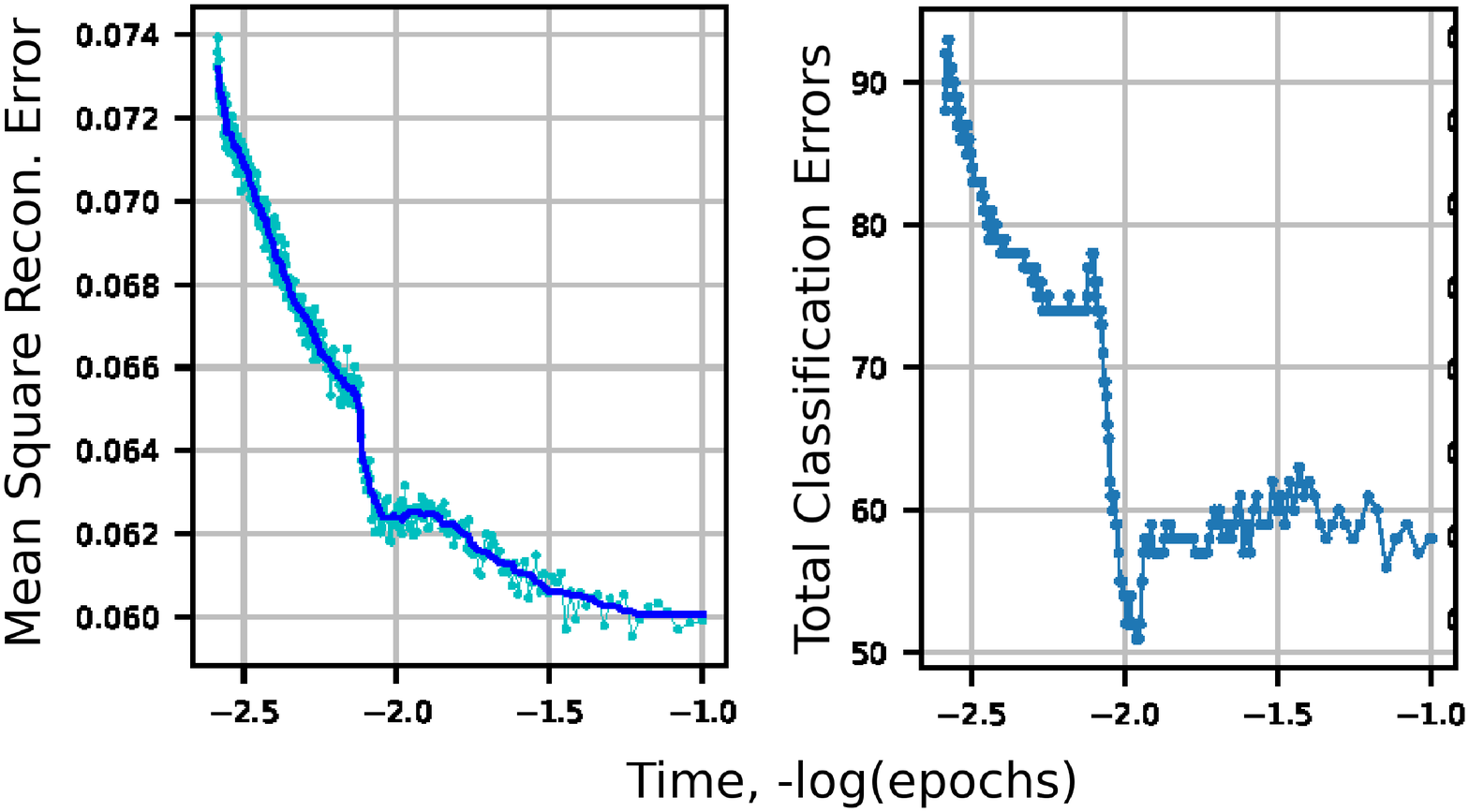}
  \caption{Training profile for the up-down algorithm
where it can be seen that when enabling TCA, both reconstruction error
and validation classfier errors decrease suddenly.
The X-axis is minus the log of the number epochs in the past.
Errors are on 1500 validation samples.}
  \label{L2updnprof}
  \end{center}
\end{figure}

\subsection{DBN performance}
The output of the first layer (using TCA) was applied to the second layer, with 
one-hot encoded labels injected, forming a DBN.
We then trained the second layer using contrastive divergence (CD) 
with three Gibbs iterations and an added term of direct free energy (FE) cost function as proposed in 
\cite{Bag2021ITG}.
Finally, the entire network was fine-tuned using the up-down
algorithm, which is an extension of CD to the entire
deep belief network \cite{HintonDeep06}.
The TCA was initialized so that it has a characteristic similar to the
base activation. Then at some point, we enabled TCA training.
In Figure \ref{L2updnprof}, it can be see at X-axis -2.1, that
TCA training was enabled, resulting in a sudden improvement
of both reconstruction error number of and classifier errors measured
on separate validation data.

%
\section{TCA for Projected Belief Network and Auto-Encoders}
\subsection{Description}
The projected belief network (PBN) is a generative network that is based on 
PDF estimation, a direct extension of (\ref{pdx2}), (\ref{pdx3})
to dimension-reducing transformations \cite{BagKayInfo2022},
so it is the ideal paradigm to test the concepts of TCA.
The PBN is based on the idea of back-projection through
a given feed-forward neural network (FFNN), 
a way to reconstruct or re-sample the input data based on the network output \cite{BagPBN}.
There are both stochastic and deterministic versions of the PBN \cite{BagPBNEUSIPCO2019}.
In the stochastic PBN, a tractable likelihood function (LF)
is computed for the FFNN, and inserting a TCA into the FFNN
applies a term to the  LF corresponding to the derivative of the TCA,
which is a mixture distribution.  The deterministic PBN (D-PBN) operates similarly, but is trained not
to maximize the LF, but to maximize the conditional LF 
(given the network output), which is a probabilistic
measure of the ability to reconstruct the input data.
The D-PBN can be seen as an auto-encoder (AEC), so we will compare
it with standard auto-encoders.
We used the same data as in Section \ref{ddesc}.


\subsection{Network}
\begin{figure}[h!]
  \begin{center}
    \includegraphics[width=3.5in]{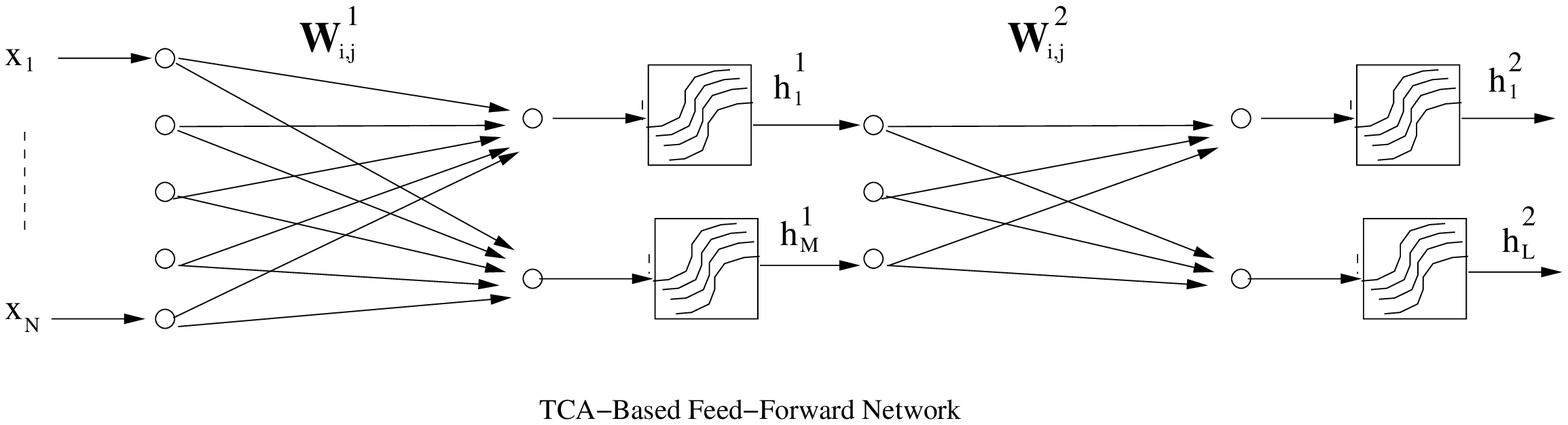}
  \caption{Illustration of a two-layer feed-forward network based on TCAs.}
  \label{pbn_nnl}
  \end{center}
\end{figure}

The network which is illustrated in Figure \ref{pbn_nnl} 
had two dense perceptron layers with 32 and 8 neurons,
respectively, and TCAs.  The base non-linearity for the TCA was TED.

\subsection{Results}
We trained the network as an AEC, a VAE, and as a D-PBN using PBN Toolkit \cite{PBNTk}.
Note that in the VAE, a TCA is not used in the the output layer,
because the output layer in a VAE has a special form.
The output TCA is also not used in the D-PBN,
since back projection starts with the output of the last linear transformation.
In all cases, we trained to convergence with TCA training disabled,
that is with the equivalent of a simple AF, then again with
TCA training enabled. We report mean square error (MSE)
on training and test data in Table \ref{tab1a}.
\begin{table}
\begin{center}
 \begin{tabular}{|l|l|l|l|}
\hline
Algorithm & TCA & MSE(train) & MSE(test)\\
 \hline
AEC & No & .02024 & .02273\\
 \hline
AEC & Yes & .01884 &  .02403\\
 \hline
VAE & No & .02220 & .02509\\
 \hline
VAE & Yes & .01835 &  .02179\\
 \hline
D-PBN & No & .01917 & .01955\\
 \hline
D-PBN & Yes & .01790 & {\bf .01790}\\
 \hline
\end{tabular}
\end{center}
\caption{Mean square reconstruction error for various auto-encoders.}
\label{tab1a}
\end{table}
We may make a number of conclusions from the table. First, using TCA
significantly improves performance in all cases 
(compare ``Yes" rows to `No" rows). Second,
the D-PBN has not only best performance, but it generalizes
much better than conventional auto-encoders,  
a feature of D-PBN that we have
reported prevously \cite{BagPBNEUSIPCO2019}.
In this case, there was almost no measureable difference between
training and test data.
As we explained, both the VAE and D-PBN do not use the final TCA,
so the performance difference hinges only on the TCA at the output of the
first layer. Despite this, a significant improvement is seen.

\subsection{TCA vs Added Network Layers}
The performance improvements of TCA in a standard feed-forward or plain 
auto-encoder can be attributed to the increased parameter count over standard
activation functions, but a TCA achieves this with far fewer parameters than adding layers.
Furthermore, in RBMs and DBNs, using TCAs creates novel generative models
with stochastic units based on finite mixture distributions, something that
cannot be achieved by adding network layers. Using TCAs, it is seen that
RBMs and DBNs have significantly better performance. 

\section{Conclusions}
In this paper, we have introduced trained compound activations
(TCAs). We justified their use based on PDF estimation
and removal of modalities. We have derived novel restricted Boltzmann machines
(RBMs) based on TCAs, and have demonstrated convincing
improvements for TCAs in experiments using stacked RBMs, deep belief networks (DBNs),
auto-encoderes and deterministic projected belief networks (D-PBNs). 
All experiments were implemented using PBN Toolkit \cite{PBNTk}.
All data, software, and instructions to repeat the results in this paper are archived at \cite{PBNTk}.

\bibliographystyle{ieeetr}
\bibliography{ppt}
\end{document}

When trained using (\ref{mleq}), the TCA approximates
the marginal cumulative distribution (histogram) of the
each input data dimension. Figure \ref{mn136hist}
illustrates this property for a single pixel of
the reduced MNIST data. The data was first
pre-processed with the inverse sigmoid function to map
it to the real line with most data values in the range $[-10,10]$.
\begin{figure}[h!]
  \begin{center}
    \includegraphics[width=3.5in]{mn136hist.eps}
  \caption{Illustration of histogram-approximation property of TCAs.
Top: histogram of input pixel data. Bottom: TCA transfer
function after training for the same pixel.}
  \label{mn136hist}
  \end{center}
\end{figure}

\subsection{First layer Results}
In the first experiments, used just the first layer.
We trained the first layer using a restricted Boltzmann machine (RBM).
Figure \ref{mn1train} shows the training profile, the graph of
reconstruction error as a function of training epoch.
A distinct change is seen when the training of the TCA parametrs was enabled.
\begin{figure}[h!]
  \begin{center}
    \includegraphics[width=3.5in]{mn1train.eps}
  \caption{first layer training profile.}
  \label{mn1train}
  \end{center}
\end{figure}
Figure \ref{mn1} reconstruction of samples
made before the TCA was trained, and after TCA training was enabled.
\begin{figure}[h!]
  \begin{center}
    \includegraphics[width=3.5in]{mn1.eps}
  \caption{Reconstruction from first layer.}
  \label{mn1}
  \end{center}
\end{figure}

\subsection{Results}
Mean square error is shown in Table \ref{tab1a}.
\begin{table}
\begin{center}
 \begin{tabular}{|l|l|l|l|}
\hline
Algorithm & TCA & MSE & LF\\
 \hline
D-PBN & No & 2.06 & \\
 \hline
D-PBN & Yes & 1.96 & \\
 \hline
\end{tabular}
\end{center}
\caption{Mean square reconstruction error vor various auto-encoders.}
\label{tab1a}
\end{table}

\bibliographystyle{ieeetr}
\bibliography{ppt}
\end{document}

Activation functions are oft
Much has been published on the comparison of generative and discriminative classifiers.
The widespread view is that discriminative classifiers
generalize better when sufficient labeled training data is available \cite{Lasserre06}.
Despite their success,  it has been recognized that
discriminative methods have flaws, vividly demonstrated by
{\it adversarial sampling} \cite{MayerAdvSamp}, a technique in which
small, almost imperceptible changes to the input data cause
false classifications.  Because generative classifiers are based on a model of the
underlying data distribution, they are less succeptible to adversarial sampling
and can complement discriminative classifiers.
There are a large number of methods that seek to combine generative and discriminative
classifiers \cite{jaakkola98exploiting,raina03classification,fng01,Fujino05,Holub08,Bosch08,Lasserre06},
or to combine discriminative and generative training \cite{Lasserre06,Minka05,BishopGenDisc}.
The weakness of generative classifiers stems from the need to estimate the data distribution,
a very difficult task that is unecessary when just classifying between known
data classes \cite{Vapnik99}.  In modeling complex data generation processes found in real-world
data, traditional probability density function (PDF) estimators such as kernel mixtures and hidden Markov models
do not suffice.   Deep layered generative networks (DLGNs) can model complex generative
processes, but the data distribution, also called likelihood function (LF) is intractable,
complicating training and inference.
Reason: the hidden variables are jointly distributed with the input data and must be integrated out.
Such networks need to be trained using surrogate cost functions such as contrastive divergence 
to train restricted Boltzmann machines \cite{WellingHinton04,HintonDeep06}, and Kullback
Leibler divergence to train variational auto-encoder (VAE) \cite{pmlr-v32-rezende14},
or an adversarial discriminative network to train generative adversarial networks 
(GAN)  \cite{GoodfellowGAN2014}.  Using DLGNs as classifiers is problematic not just
because of the intractability of the LF, but because the performance of generative
models in general lags behind discriminative classifiers. 
Significant performance improvements could potentially be made by combining
DLGNs with deep discriminative networks. But, in 
order to see a benefit by combining, all classifiers need to have good performance. 
Therefore, having a DLGN classifier with comparable 
performance to deep discriminative network would be greatly desireable.

\bibliographystyle{ieeetr}
\bibliography{ppt}
In summary, there is a need for a DLGN with tractable LF  
that can be combined with discriminative approaches.
The newly introduced layered generative network called projected belief network (PBN) 
stands out as a potentially better choice to achieve these goals.
The PBN is a DLGN, so can model complex generative processes,
but stands out from all other DLGNs.  The tractable LF allows
direct gradient training and enables detection of out-of-set samples 
(outliers that are outside of the set of training classes).
Because the PBN is based on a feed-forward neural network (FF-NN), it can share 
an embodiment with a discriminative classifier (i.e. it is a single network
that is both a complete generative model and a discriminative classifier),
so is a more direct way to introduce the
advantages of generative models into a discriminative classifier, or vice-versa.

\subsection{Main Idea}
The PBN is based on a feed-forward neural network (FF-NN).
Figure \ref{asy0} shows a simple 3-layer FF-NN.
\begin{figure}[h!]
  \begin{center}
    \includegraphics[width=3.5in]{asy0.eps}
  \caption{A feed-forward neural network (FF-NN).  This FF-NN can be a discriminative classifier
if $\lambda_4$ is the {\it softmax} function and the output
box is the cross-entropy cost function.  It can also be a generative model if viewed as  PBN
 and  the output box is the output prior distribution  $g(\bfx_4)$.}
  \label{asy0}
  \end{center}
\end{figure}
Each layer $l$ consists of a linear transformation (represented by matrix ${\bf W}_l$),
a bias $\bfb_l$ and an activation function $\lambda_{l+1}(\;)$.
The linear transformation can be fully-connected or convolutional,
but must have total output dimension equal to or lower than the input dimension.
This network can serve as a traditional classifier network if the output layer dimension is 
equal to the number of classes (and the final activation function  is {\it softmax}).
On the other hand, it can also be viewed as a projected belief network (PBN) \cite{BagPBN,BagEusipcoPBN}
whose properties are reviewed below.

\subsection{Mathematical Foundation}
Suppose we are given a fixed dimension-reducing transformation mapping
high-dimensional input data $\bfx\in \mathbb{R}^N$ to 
a lower-dimensional feature $\bfz\in \mathbb{R}^M$, $M<N$, denoted by $\bfz = T(\bfx)$.
If the probability density function (PDF) of the feature, denoted by $g(\bfz)$,
is estimated or specified, then we may ask the question ``given
$g(\bfz)$ and $T(\bfx)$, what is a good estimate of the PDF of $\bfx$?". 
The method of maximum entropy (MaxEnt) PDF projection \cite{Bag_info}
finds a unique PDF defined on $\mathbb{R}^N$ with highest
entropy among all PDFs consistent with $T(\bfx)$ and $g(\bfz)$. 
When PDF projection is applied layer-wise to a feed-forward neural network (identified with $T(\bfx)$) then 
a projected belief network (PBN) results \cite{BagPBN,BagEusipcoPBN}.
The LF for the network in Figure \ref{asy0} is given by (see \cite{BagEusipcoPBN}) 
\beq
\begin{array}{l}
p_p(\bfx_1; T, g) = 
\frac{1}{\ds \epsilon} \; \frac{\ds p_1(\bfx_1)}{\ds p_1(\bfz_1)} \;  |{\bf J}_{\bfz_1 \bfx_2}|  \\  
  \;\;\;\;\;\;\;\;\; \cdot \; \frac{\ds p_2(\bfx_2)}{\ds p_2(\bfz_2)} \; |{\bf J}_{\bfz_2 \bfx_3}| \;
	\frac{\ds p_3(\bfx_3)}{\ds p_3(\bfz_3)} \;  |{\bf J}_{\bfz_3 \bfx_4}| \; g(\bfx_4),
\end{array}
\label{cr1a}
\eeq
where $p_l(\bfx_l)$ is the assumed prior distribution for the input to layer $l$, 
 $p_l(\bfz_l)$ is the distribution of $\bfz_l$ under the assumption that
$\bfx_l$ is distributed according to $p_l(\bfx_l)$, 
$|{\bf J}_{\bfz_l \bfx_{l+1}}|$ is the determinant of the Jacobian (matrix of gradients)
of the invertible transformation mapping $\bfz_l\rightarrow \bfx_{l+1}$, and where $g(\bfx_{L+1})$ is the assumed prior for the output of a network.
The constant $\epsilon$ is the {\it sampling efficiency} discussed below.

\subsection{PBN layers and MaxEnt Priors}
The properties of PBN layer depend greatly on the assumed prior distribution $p_l(\bfx_l)$,
which is selected using the principle of Maximum Entropy (MaxEnt) and depends on the assumed input data 
range for the layer \cite{BagIcasspPBN}.
Consider a generic PBN layer with input dimension $N$.
There are three canonical input data ranges, {\it unlimited} denoted by $\mathbb{R}^N$,
{\it positive quadrant} where $0 < x_i$ denoted by $\mathbb{P}^N$,
and the {\it unit hypercube} where $0 < x_i < 1$ denoted by $\mathbb{P}^N$.
The MaxEnt priors for these data ranges are given in Table \ref{tab1v}.
For each data range and prior, there is a prescribed
input non-linearity (activation function) $\lambda(\;)$ which is also given in the table and should be applied at the output
of the previous layer.
The TG and TED activations resemble {\it softplus} and {\it sigmoid}, respectively \cite{BagIcasspPBN}.
Dimension-preserving layers are also possible ($M=N$) and are analyzed using the
determinant of the Jacobian matrix of the transformation.
\begin{table}
\begin{center}
 \begin{tabular}{|l|l|l|}
\hline
	 ${\cal X}$ &  MaxEnt Prior $p_0(\bfx)$  & $\lambda(\alpha)$ \\
 \hline
	 $\mathbb{R}^N$   & $\prod_{i=1}^N {\cal N}(x_i)$  (Gaussian) & $\alpha$  (Linear)\\
 \hline
	 $\mathbb{P}^N$   & $\prod_{i=1}^N 2 {\cal N}(x_i), \;\; 0<x_i $ (TG) & $\alpha + \frac{{\cal N}(\alpha)}{\Phi(\alpha)}$ (TG)\\
 \hline
	 $\mathbb{U}^N$   & $1, \;\; 0<x_i<1$  (Uniform) & $\frac{e^{\alpha}}{e^{\alpha} - 1}-\frac{1}{\alpha}$ (TED)  \\
 \hline
\end{tabular}
\end{center}
\caption{MaxEnt priors and activation functions as a function of input data range.
	TG=``Trunc. Gauss.". TED=``Trunc. Expon. Distr".
${\cal N}\left(x\right) \defined \frac{e^{-x^2/2}}{\sqrt{2\pi}}$ and $\Phi\left( x\right)  \defined \int_{-\infty}^x {\cal N}\left(x\right).$
}
\label{tab1v}
\end{table}

%
The sampling efficiency $\epsilon$ is less than 1 if the
output range of a layer is not identical to the assumed
input range of the next layer resulting in {\it subspace mismatch}.
However, $\epsilon$ is driven closer to 1 as the network trains
\cite{BagEusipcoPBN}, so can be ignored (i.e. assumed to be 1.0)  for all practical purposes.  
It is also possible to create PBNs with sampling efficiency identically equal to 1 by
using layers with Gaussian prior and 
dimension-preserving layers ($M=N$), both of which have no subspace mismatch.
The PBN is trained by maximizing the mean of the log of (\ref{cr1a})
using stochastic gradient ascent.  

\section{Technical Approach}
\subsection{Output Non-Linearity and Prior}
In order to create a PBN that is also a discriminative classifier,
a label-dependent output non-linearity and output prior are required.
A simple approach would be to apply a label-dependent level-shift
to the output variables, then assume a zero-mean standard normal output prior.
Specifically, one applies a level-shifting function $\lambda(z_i) = z_i-l_i$, $1\leq i \leq M$,
where $z_i$ is the network output (prior to activation function),
where $M$ is the number of data classes and the dimension of the network output, and 
${\bf l}=[l_1,l_2 \ldots l_M]$ is the label signal, a 
shifted one-hot encoding of the ground-truth label, with elements taking values of
$-L$ or $L$. Then, training with the simple Gaussian prior $g(\bfx_{L+1}) = {\cal N}(\bfx_{L+1}),$
where ${\cal N}(\bfx)=-\frac{M}{2}\log(2\pi)-\frac{1}{2} \bfx^\prime \bfx$ 
encourages the network output to agree with the label signal.
But, this approach makes little penalty for classification errors.  The function
$$
  \lambda(z_i) = z_i+C[\sigma(3 z_i)-.5]-l_i*(L+C/2)/L,
$$
where $\sigma(\;)$ is the sigmoid function and $C$ is a large constant,
produces a dynamic level shift that greatly increases if the
network output has the wrong sign (compared to the label).
When combined with the standard normal prior, it encourages 
Gaussian modes at -$L$ and +$L$, but imposes a very large penalty for class errors.
The degree of discriminative training can be varied by changing $C$.  

\subsection{MaxEnt Reconstruction and Synthesis}
\label{reconsec}
We now investigate a distinctly generative property of the
PBN : visible data reconstruction from hidden variables.
Input data can be randomly synthesized or
reconstructed from the output of any layer of the FF-NN.
Unlike other generative networks, the PBN is not an
explicit generative network, it operates implicitly 
by ``backing up" through a FF-NN.
In each layer,  the PBN selects a sample from the set
$${\cal M}(\bfz) = \{ \bfx : {\bf W}^\prime \bfz=\bfx\},$$
which is the set of samples $\bfx$ that ``could have" produced
$\bfz$.  A sample is selected from ${\cal M}(\bfz)$
with probability density proportional to the prior distribution $p_0(\bfx)$. 
When $p_0(\bfx)$ is the uniform
distribution, this is called uniform manifold sampling (UMS)  \cite{BagUMS}.
Sampling requires a type of Markov chain Monte-Carlo (MCMC) \cite{BagUMS}.
Deterministic data generation is also possible if instead
of randomly selecting a sample in ${\cal M}(\bfz)$, we
select the mean (the conditional mean given $\bfz$) ,
denoted by $\hat{\bfx}|\bfz = {\mathbb E}(\bfx|\bfz).$
This can be found in closed form for
a range of MaxEnt priors \cite{BagIcasspPBN,BagEusipcoPBN,BagUMS}
and  is given by $\hat{\bfx}|\bfz =  \lambda({\bf W} \hat{\bfh}),$
  where $\lambda(\;)$ is given in Table \ref{tab1v} and
$\hat{\bfh}$ is the solution of the equation 
\beq
{\bf W}^\prime \lambda\left({\bf W} \bfh \right)=\bfz.
\label{hsola}
\eeq
This solution is guaranteed to exist as long as $\bfx$ is in the support $p_0(\bfx)$
and is also the saddle-point for the saddle-point approximation to $p_0(\bfz)$  \cite{BagIcasspPBN}.
For the simplest case of Gaussian MaxEnt prior, the activation function is linear, $\lambda(\alpha)=\alpha$,
and the reconstruction is by least-squares, $\hat{\bfx}|\bfz = {\bf W} \left({\bf W}^\prime {\bf W}\right)^{-1}\bfz.$

Starting at any layer output, one can proceed in the backward direction
up the network, increasing the dimension, until the visible data
is reconstructed.  There are two possible reconstruction methods, (a) 
random sampling in ${\cal M}(\bfz)$ by MCMC, and (b) deterministically
selecting the conditional mean   $\hat{\bfx}|\bfz$.
When subspace mismatch occurs,
the reconstruction chain could fail. The rate of success is the
{\it sampling efficiency} discussed above, and is different for
stochastic and deterministic reconstruction.
Generally, a trained network has a deterministic sampling efficiency of 1
(failure is rare or non-existent).
Reconstructing from dimension-preserving layers involves just a matrix inversion,
so has a sampling efficiency of 1.
Layers with Gaussian input assumption also have a sampling efficiency of 1.
Deep networks can be constructed using these two layer types to obtain
deep PBNs with sampling efficiency of 1.

When only determintstic reconstruction is used, the result is a deterministic
PBN \cite{BagEusipcoPBN}, a type of auto-encoder where the reconstruction
network that is defined by the analyis network.

\subsection{PBN Properties}
The PBN differs significantly from other methods
of combining the roles of generative and discriminative networks because the
discriminative influence is added into the output prior
and does not disturb the ``purity" of the generative network.
There is no compromise between generative and
discriminative training or structure, they are both 
contained in one network and one cost function.

When reconstructing visible data from hidden variables, 
the synthesized data, when applied to the feed-forward
network, produces exactly the same hidden variables as were
created during the generation process. This property of hidden variable recovery
is unique to the PBN.  

During training, when the discriminative cost function is ``satisfied" (the training data is almost 
completely separated), then the generative cost dominates, so the network becomes the best possible
PBN that at the same time separates the data.  This can be seen as a generative regularization effect.

\section{Classification of Spectrograms of Words Commands}
\subsection{Data set}
The data was selected to be at the same time relevant, realistic,
and challenging.  We selected a subset of the Google speech commands data \cite{GoogleKW},
choosing three pairs of difficult to distinguish words: ``three, tree",
``no, go", and ``bird, bed", sampled at 16 kHz and segmented into 
into 48 ms Hanning-weighted windows shifted by 16 ms.  
We used log-MEL band energy features with 20 MEL-spaced
frequency bands and 45 time steps, representing a frequency span of 8 kHz and a time span of 0.72 seconds.
The input dimension was therefore $N=45\times 20=900.$
From each of the six classes, we selected 500 training samples, 150 validation samples, at random.
The remaining samples were used to test, averaging about 1500 per class or about a total of 10000
testing samples.

\subsection{Network}
A separate network was trained on each word pair.
The networks had $L=5$ layers.  The first layer was convolutional with
9 ($21\times 17$) convolutional kernels using ``same" border mode and
$5\times 4$ downsampling (not pooled, just down-sampled), thus producing
9 ($9\times 5$) output feature maps, or a total output dimension of 405.
The second layer was convolutional with
24 ($5\times 3$) convolutional kernels using ``same" border mode and
$2\times 2$ downsampling, thus producing
24 ($3\times 2$) output feature maps, or a total output dimension of 144.
The remaining two layers were fully-connected with 64, and 24 neurons.
The output layer had 2 neurons, matching the number of classes. 
Note that we sought to reduce the dimension in each layer by at least a factor of 2.
The layer output activation functions were linear, linear, TG, TG, and linear 
(See Table \ref{tab1v}).

\subsection{Classification Results}
For a classifier benchmark, we trained a conventional CNN classifier network
on each class pair.  Each network consisted of
seven layers, three convolutional and four dense layers.
The convolutional layers had kernel shapes of $(11\times 5)$, $(7\times 5)$, and $(3\times 3)$,
max-pooling of  $(5\times 2)$, $(3\times 2)$, and $(1\times 1)$,
with 64, 32, and 48 kernels, respectively.
Soft-plus activation and ``same" convolutional padding was used.
The dense layers had  256, 128, 32 , and 2 neurons.
Dropout and batch normalization were used with ADAM optimization.
Classification accuracy for the CNN is given in Table \ref{tab1}.

The PBN networks were first initialized with random weights and 
trained as a standard discriminative deep neural network (DNN)
with dropout and L-2 regularization.
No data augmentation (such as random shifting) was used.
Classification accuracy for the initial PBNs is given in Table \ref{tab1} as `PBN(DNN)".
\begin{table}[htb]
\begin{center}
 \begin{tabular}{|l|l|l|l|}
\hline
 & \multicolumn{1}{|c|}{"three-tree"} &  \multicolumn{1}{|c|}{"no-go"}  & \multicolumn{1}{|c|}{"bird-bed"} \\
 \hline
         CNN &    0.923 &  0.924  & 0.965 \\

 \hline
	 PBN(DNN) &  0.873 &  0.859 &  0.960 \\
 \hline
	 PBN &  0.886 &  0.863 &  0.960 \\
 \hline
	 PBN+CNN &  {\bf 0.925} &  {\bf .926} &  {\bf 0.971} \\
 \hline
\end{tabular}
\end{center}
	\caption{Classification accuracy for the three class pairs.}
	\label{tab1}
\end{table}
The initialized PBN were then trained 
as a PBN by maximizing the mean likelihood function (\ref{cr1a})
with output prior distribution parameter  $C=200$ and L2 regularization.
The classification results for the PBN on the three class pairs are shown
in Table \ref{tab1} where they can be compared with the initial
DNN-trained networks ``PBN(DNN)". Note that
the PBN has about the same accuracy as ``PBN(DNN)", with slightly higher accuracy for two class pairs.
This demonstrates that the PBN training does not seem to impact the classifcation
performance of a network, and may even help.
Training a classifier network as a PBN can be regarded as a form of regularization.

\subsection{Reconstruction Results}
It has been established that the PBN has lost little in terms of
classification performance wwhen compared to the
initial regularized discriminative networks. It will now be determined
what has been gained in terms of generative power.
The first thing that comes to mind is the reconstruction of visible data
from the hidden variables.  Using the method of Section \ref{reconsec},
we reconstructed data from the hidden variables
of the first and second layer of the initial PBN ``PBN(DNN)", with dimensions
405 and 144, respectively.  Results are shown in Figure \ref{dnnrecon}.
\begin{figure}[h!]
  \begin{center}
    \includegraphics[width=3.5in,height=1.0in]{cdbn71_45_dnn_recon.eps}
  \caption{Samples of spoken word commands 
	``three" and ``tree" and reconstruction using the discriminatively-trained network.
	  From top: original samples, first-layer reconstructions, 
	  second-layer reconstructions.  }
  \label{dnnrecon}
  \end{center}
\end{figure}
Little resemblance can be seen despite the high dimension
of the hidden variables.  This is how the network sees the data through the hidden variables.
The noisy images, when used as input data will produce exactly
the same hidden variables at the given layer as the
original input sample, a disturbing fact that vividly illustrates one of the
problems with discriminative networks.

The reconstruction experiment was repeated for the trained PBN.
Results are shown in Figure \ref{pbnrecon}. 
\begin{figure}[h!]
  \begin{center}
    \includegraphics[width=3.5in]{cdbn71_45_pbn_recon0.eps}
  \caption{Samples of speech commands ``three, three" reconstructed using PBN. From top: original spctrograms, then the same reconsructed 
from output of first through fourth layers, with hidden variable dimensions of 405, 144, 64, and 24,
	  respectively.}
  \label{pbnrecon}
  \end{center}
\end{figure}
This time, reconstruction was attempted from deep within the network.
The reconstructions had excellent quality, but gradually decreasing sharpness.  
Note that this network was not trained for lower reconstruction error, but instead to maximize (\ref{cr1a}).
The reconstruction power of the network comes as a side-effect and can be tapped 
into anywhere in the network.  

\subsection{Classifying between class pairs}
A second exercise in ``generative capability" is 
the classification between class pairs using models trained separately
on just one pair.  This demonstrates the ability to recognize out-of set events.
To classify between pairs, visible data reconstruction error was calculated
based on the 24-dimensional output of the fourth layer (not using the output layer),
and the model giving least error was chosen.
Figure \ref{sixclass} shows the classifier statistic (negative log of mean square reconstruction error).
The inter-pair classification accuracy was 87.9\%, which is good considering
the number of mal-formed events in the data base and
that the models were separately trained, without access to data
of the competing class pairs.  
\begin{figure}[h!]
  \begin{center}
    \includegraphics[width=3.5in,height=1.0in]{classpairs.eps}
	  \caption{Classifier statistic (negative log of 
	  mean square reconstruction error) for classifying between class pairs
	  based on reconstruction error.}
  \label{sixclass}
  \end{center}
\end{figure}

\subsection{Combination with CNN}
We have postulated above that having a generative classifier
with comparable performance to a discriminative one would
allow for performance gains when combining them.
To demonstrate this, the PBN was combined with the  CNN benchmark classifier
described above.  In Figure \ref{classcomb}, combined classifier error in percent is shown
as a function of additive combination weight for each class pair.
As might be expected, the class-pair in which the generative and
discriminative performance are the most similar (see Table \ref{tab1})
shows the most improvement.
\begin{figure}[h!]
  \begin{center}
    \includegraphics[width=3.5in,height=1.5in]{classcomb.eps}
  \caption{Classifier combination results for each class pair as a function
	  of linear combination weight. Performance at the
	  far right of each graph corresponds to CNN only
	  and far left to PBN.}
  \label{classcomb}
  \end{center}
\end{figure}

\subsection{Random Synthesis}
As a final demonstration of generative power, we synthesized entirely random
events by starting with random data equal in dimension to the PBN
output layer, in this case dimension-2.
Data was synthesized at the point prior to the output
activation function using Gaussian random variables.
Results are shown in Figure \ref{syn45} for the class
pair ``three" and ``tree".
\begin{figure}[h!]
  \begin{center}
    \includegraphics[width=3.5in]{syn45.eps}
  \caption{Top: ten training samples randomly selected from
	   ``three" and ``tree" spoken word commands. Bottom: randomly synthesized
	   data from trained PBN. There is no relationship to the
	   selected training samples on top.}
  \label{syn45}
  \end{center}
\end{figure}
The synthetic samples appear realistic and are diverse,
showing variations in time shift, dilation, and other qualities.
This means that the PBN has indeed learned much about the
data generation process.

\subsection{Implementation and Applications }
The PBN was implemented in Python using Theano 
symbolic expression compiler \cite{Theano}.
The primary computational challenge is the solution
of a symmetric linear system with dimension $M\times M$,
where $M$ is the total output dimension of a layer.
This must be solved for each iteration in the solution
of (\ref{hsola}).  This was parallelized on the GPU, one processor
for sample in a mini-batch.  The computational time for an epoch was 1.1 seconds.
This was only about an order of magnitude slower than training the DNN.
All results were obtained using PBN Toolkit \footnote{http://class-specific.com/pbntk. A copy
of the data is also available at this link}.

\section{Conclusions }
In this paper, a projected belief network (PBN), which is a purely
generative layered network,  was trained as a
generative-discriminative classifier.  This was achieved using a label-dependent prior for the output features.
Since the PBN is based on a feed-forward neural network (FF-NN),  it can share
an embodiment with a discriminative deep neural network (DNN). 
Through the parameter $C$, the network can be trained with varying amount of discriminative influence.
When reconstructing visible data from the hidden variables, it was shown that the
the same netrork, trained discriminatively, had very poor ability to reconstruct, even from initial layers,
whereas whe the network was trained as a PBN, the reconstruction greatly improved.
The PBN classifier had comparable classification performance
to the discriminatively-trained network, yet provided generative power
from three standpoints: visible data reconstruction from hidden variables,
random data synthesis, and classification of out-of set samples.
It was also shown to improve upon a conventional CNN when
additively combined.

\bibliographystyle{ieeetr}
\bibliography{ppt}
\end{document}

\subsection{Background and Motivation }
Discriminative neural networks have dominated machine learning for decades.
The performance of generative networks lags behind 
because they need to model the generative process underlying the data, a much harder
task than discrimination \cite{Vapnik99}. Yet, interest in generative models persists
because a model of the underlying process is useful, as exemplified by variational
autoencoders (VAE) \cite{pmlr-v32-rezende14}, and generative adversarial network \cite{GoodfellowGAN2014}
(GAN) which have sparked considerable interest.
While the generative task is harder, given time and effort, 
generative models can perform as well as classifiers as
their discriminative counterparts. 
For example, when Hinton's deep belief network (DBN) 
was published, the DBN worked better than comparable fully-connected 
(non-convolutional) feed-forward networks \cite{HintonDeep06}.
While training algorithms have been developed for VAE and DBN,
the likelihood functions (LF) are not available in closed-form, so need to be approximated,
using stochastic variational methods in the case of VAE \cite{pmlr-v32-rezende14},
or Monte Carlo approximations in the case of DBN \cite{SalakhutdinovDBN}.
%
%
The projected belief network (PBN) is 
a new type of generative network with tractable 
LF that generates data layer-wise from hidden variables similar to a 
 deep latent Gaussian model (DLGM).  But, in contrast to other
generative models, the PBN  is related to a feed-forward neural 
network (FF-NN)  by a duality relationship \cite{BagPBN}.  
The dual FF-NN, which is here called dual analysis network (DAN), 
exactly recovers the hidden variables
of the PBNs data generation process.  
%
%
With tractable LF, the PBN has the potential to enable a new
class of generative models and algorithms.

\subsection{Main Idea}
The projected belief network (PBN) was previously introduced as a dual counterpart
to a feed-forward neural network (FF-NN) \cite{BagPBN}.
The PBN is derived from a FF-NN by asking the following question: {\it knowing
the FF-NN and the distribution of the output variables (features) of the FF-NN,
what is the  maximum entropy (MaxEnt) distribution of the visible 
data consistent with the given features distribution?}
The PBN is the generative network that implements this MaxEnt distribution \cite{BagPBN}.
Not surprisingly, the PBN uses the same network weights as the FF-NN
from which it is derived, and employs a special 
``activation" function that gives it its unique properties.
A deterministic version of the PBN is created if instead of
generating random data in each layer, the conditional
mean is propagated.  The deterministic PBN is the complementary
network to the DAN and combined with the DAN forms a new
type of auto-encoder.

\subsection{Paper Contributions}
The PBN has been previously introduced \cite{BagPBN}. Novel contributions of this paper include
(a) experimental results comparing PBN with 
other models as a function of data dimension,
(b) the detailed description of a multi-layer PBN,
(c) the treatment of the issue of sampling efficiency,
(d) the conceptual comparison of PBN with the VAE,
and (e) the description of a deterministic PBN
and its application as an auto-encoder,
and experiments showing significant improvements
over a conventional auto-encoder of the same structure.

\section{Projected Belief Networks (PBN)}
\subsection{PBN Exact Form}
Figure \ref{pbn_multi} illustrates a two-layer PBN in its exact, asymptotic,
and deterministic forms.  It can be easily extended to more layers.
\begin{figure}[h!]
  \begin{center}
    \includegraphics[width=3.5in]{pbn_multi_b.eps}
  \caption{A 2-layer PBN in three forms,
exact, asymptotic, and deterministic, and the
corresponding dual analysis network (DAN).}
  \label{pbn_multi}
  \end{center}
\end{figure}
Near the bottom of the figure is the dual analysis network (DAN), a conventional
feed-forward network employing an activation function
$\lambda_n(\;)$ in layer $n$.
Optionally, an energy statistic (ES), denoted by $e=t(\bfx)$ is extracted from the input
of each layer.
The figure illustrates both data generation by different forms of the
PBN (left to right) and feature extraction by the DAN (right to left).
Data generation originates by a feature generating distribution
$g(\bfz_2)$, then continues layer by layer.  In layer $n$ of the exact form of the PBN, (top), the
activation function and bias (if used) are inverted, and the
feature $\bfz_n$ is presented to the ``UMS" block in which a 
sample $\bfx$ is drawn randomly from the set ${\cal M}_n(\bfz_n,e_n)$ defined by
\beq
   {\cal M}_n(\bfz_n,e_n) = \{ \bfx : {\bf W}_n^\prime \bfx = \bfz_n, \; t_n(\bfx)=e_n, \;\;\; \bfx \in {\cal X}_n \},
   \label{manifze}
\eeq
where ${\cal X}_n$ is the input range of layer $n$ and $e_n=t_n(\bfx)$ is the optional ES.
The sample $\bfx$ must be drawn with uniform distribution,
so that no member of ${\cal M}_n(\bfz_n,e_n)$ is more likely to be drawn than any other.
The sampling procedure is therefore called uniform manifold sampling (UMS) \cite{BagUMS}.

By the definition of UMS, the DAN will exactly recover the variables $\bfz_2$, $\bfz_1$. 
When the PDF of $\bfz_2$ is known, denoted
by $g(\bfz_2)$, then the PBN generates samples the PDF:
\beq
p_p(\bfx_1; T, g) = \frac{1}{\epsilon} \; \frac{p(\bfx_1 ; H_{0,1})}{p(\bfz_1 ; H_{0,1})} \;  |{\bf J}_{\bfz_1 \bfx_2}| \; \frac{p(\bfx_2 ; H_{0,2})}{p(\bfz_2 ; H_{0,2})} 
\; g(\bfz_2),
\label{cr1a}
\eeq
where $\bfx_n$ is the input data to layer $n$ ($\bfx_1$ is the visible data), 
$T$ represents the DAN, $|{\bf J}_{\bfz_1 \bfx_2}|$ is the determinant of the
Jacobian of the 1:1 mapping from $\bfz_1$ to $\bfx_2$,
and $\epsilon$ is the sampling efficiency, to be explained below.

Notice the absence of integral signs in (\ref{cr1a}) - the distribution
does not require integrating out the hidden variables, as is necessary
in other layered generative models.  This is due to the fact that the
hidden variables of the DAN are deterministically derived from the
visible data, not jointly distributed.   Note also that in (\ref{cr1a})  
there appears a set of reference distributions, one for each layer.
The distribution $p(\bfx_n;H_{0,n})$ is the maximum entropy (MaxEnt) reference distribution
for layer $n$ and $p(\bfz_n;H_{0,n})$ is the corresponding feature distribution\footnote{$p(\bfz_n;H_{0,n})$ is the theoretical PDF
of the layer output when the layer input is distribued according to $p(\bfx_n;H_{0,n})$.}.
This reference distribution depends on ${\cal X}_n$, the data range of layer $n$ input,
which in turn depends on the activation function used in the previous layer - note
that the input (visible data) is assumed to have been created using $\lambda_1(\;)$.
We consider three data ranges: $\mathbb{R}^N$, $\mathbb{P}^N$,
and $\mathbb{U}^N$, where $N$ represents input data dimension of a generic layer,
$\mathbb{R}^N$ is the unlimited case, $\mathbb{P}^N$ is the positive quadrant ($0\leq x_i$),
and $\mathbb{U}^N$ is the unit hypercube ($0\leq x_i \leq 1$).
%
The MaxEnt reference distribution for each data range ${\cal X}$ is given in Table \ref{tab1v}.
The primary computational challenge in computing (\ref{cr1a})
is calculating the denominator terms $p(\bfz_n;H_{0,n})$.  More is provided in the 
references \cite{BagPDFProj,BagNutKay2000,Bag_info,BagUMS,BagPBN,BagEusipcoRBM}.
\begin{table}
\begin{center}
 \begin{tabular}{|l|l|l|l|l|}
\hline
${\cal X}$ &  $p(x;\alpha)$ & $\lambda(\alpha)$  & $t(\bfx)$ & $p(\bfx;H_0)$\\
 \hline
$\mathbb{R}^N$   &  ${e^{-(x-\alpha)^2/(2\sigma^2)} \over \sqrt{2 \pi \sigma^2} }$ (Gauss.) & $\alpha$  & $\sum_i x_i^2$  
& $\frac{e^{-t^2(\bfx)/2}}{(2\pi)^{-N/2}}$\\
 \hline
$\mathbb{P}^N$  &  $\alpha e^{-\alpha x}$  {\hspace{.37in}} (Expon.) & $1/\alpha$    & $\sum_i x_i$  & $e^{-t(\bfx)}$ \\
 \hline
$\mathbb{U}^N$   &   $\left(\frac{\alpha}{e^{\alpha} - 1}\right)  \; e^{\alpha x}$   {\hspace{.12in}} (TED) & $\frac{e^{\alpha}}{e^{\alpha} - 1}-\frac{1}{\alpha}$    & none & 1\\
 \hline
\end{tabular}
\end{center}
\caption{Generating distributions $p(x;\alpha)$, expected value of generating distributions $\lambda(\alpha)$,
energy statistics (ES) $t(\bfx)$,  and reference hypotheses $p(\bfx;H_0)$
for for data ranges $\mathbb{R}^N$, $\mathbb{P}^N$, and $\mathbb{U}^N$.
This table concerns a single layer and ${\bf x}$ is assumed to be the
visible data for the given layer layer with dimension $N$ and range ${\bf x}\in {\cal X}$.
}
\label{tab1v}
\end{table}


Depending on the data range (see Table \ref{tab1v}) an ES might need to be extracted from each layer input. We describe the ES for completeness,
but no ES is needed for $\mathbb{U}^N$, and for $\mathbb{P}^N$, the 
ES can be incorporated into matrix ${\bf W}_n$, eliminating the need for an explicit ES.  For more about the ES, please consult the references \cite{Bag_info,BagUMS}.

Optionally, a bias and activation function can be appended to the DAN (bottom of Figure \ref{pbn_multi}),
producing feature $\bfx_3$. In this case, the data generation process begins with the generating distribution
$g(\bfx_3)$, and the activation function and bias must be inverted.  Also, 
equation (\ref{cr1a}) must be modified by replacing $g(\bfz_2)$ with $|{\bf J}_{\bfz_2 \bfx_3}| \;g(\bfx_3).$

%

\subsection{PBN Asymptotic Form}
It has been shown that the UMS sampling process can be closely approximated by a
network layer resembling a sigmoid belief network \cite{BagUMS}.
To arrive at the asymptotic PBN (see Figure \ref{pbn_multi}), 
the UMS blocks are replaced by a nonlinear function $\bfh_n = \gamma_n^{-1}(\bfz_n)$,
 matrix multiplication $\balpha_n = {\bf W}_n \bfh_n$, then generation
from distributions $p_n(x; \alpha)$, which
are given in Table \ref{tab1v} as a function of ${\cal X}_n$.  The expected value of these distributions
(given $\alpha$) is denoted by $\lambda_n(\alpha)$, which
corresponds to the activation functions used in the DAN
at the output of layer $n-1$.  
Interestingly, for $\mathbb{U}^N$, $\lambda_n(\alpha)$
 is the mean of the truncated exponential distribution (TED),
which is similar to the sigmoid function \cite{BagUMS}.
Central to the theoretical analysis of a PBN layer
is the function $\gamma_n(\bfh_n)  =  {\bf W}_n^\prime \lambda( {\bf W}_n \bfh_n).$
To compute a layer of a PBN, this function needs to be inverted:
$\bfh_n=\gamma_n^{-1}(\bfz_n),$ which requires  
an iterative algorithm, but
might have no solution (See Section
\ref{sampeff}).

\subsection{The PBN for $\mathbb{R}^{N}$ and Relationship to VAE}
The VAE is currently a well-studied generative model \cite{Goodfellow2016,pmlr-v32-rezende14}.
The ``variational" aspect of VAE has to do with approximating and training the LF,
but the VAE is essentially an implementation of DLGM \cite{pmlr-v32-rezende14}.
Thus, both PBN and VAE are layered generative models. The main difference
is that the PBN is based on an explicit feed-forward analysis network (the DAN),
so the latent variables can be  deterministically computed from the
visible data. So, once a visible data
sample has been generated by the PBN, all the hidden variables can then be exactly
recovered by a single pass of the DAN.
The VAE on the other hand is a stochastic layered generative model,
so the latent variables of the VAE are jointly distributed
with the visible data. For this reason the LF of the VAE is only available 
as an integral over the hidden variables.
But, this distinction is moot because when looking at the asymptotic form of the PBN,
an approximation that is very good as has been demonstrated \cite{BagUMS},
we see that the PBN {\it behaves} like a traditional layered stochastic generative model.

A network layer of a DLGM is composed of an arbitrary
non-linear function followed by additive correlated noise \cite{pmlr-v32-rezende14}.
A network layer of an asymptotic PBN, on the other hand,  is composed of 
a non-linear function $\gamma_n^{-1}(\bfz)$, followed by multiplication by matrix 
${\bf W}_n$, then the generating distributions are applied
to produce the output variables.  Function $\gamma_n^{-1}(\bfz)$ and the generating distributions  
depend on the range of the layer output variable and are given in Table
\ref{tab1v}.  When $\bfx \in \mathbb{R}^{N}$, the generating distribution is
Gaussian, and is implemented by adding independent Gaussian noise\footnote{This can be
easily extended to correlated noise by introducing a matrix multiplification 
between the layers.}.  This produces a type of DLGM.
But, the Gaussian noise in an asymptotic PBN must be added
after a linear transformation, whereas for DLGM it is added after an arbitrary transformation.
It is not clear what this distinction means to the ultimate
PDF estimation capability, and can only be discovered by future experiments. 
Note also that for the DLGM, the activation function is taken to be
part of the ``arbitrary non-linear function" , whereas in the 
PBN, the activation function $\lambda_n(\;)$ is defined for the
dual DAN, which determines the function  $\gamma_n^{-1}(\bfz)$ used in the PBN.
In holding to the MaxEnt principle, 
for a given data range ${\cal X}$, the activation function 
$\lambda()$ is fixed, and therefore $\gamma_n^{-1}(\bfz)$ is fixed.
But, if one is willing to give up this MaxEnt distinction, there is 
flexibility in choosing $\lambda()$  so long
as it is invertible (for example use {\it softplus}, not {\it relu}).

In summary, both DLGM and PBN are layered
generative networks and it is not clear from the
above comparison which structure is better or more general.
It is clear, however, that the PBN under special conditions
(i.e. for ${\cal X}=\mathbb{R}^N$) approximates a type of DLGM and has a closed-form LF
 which is especially efficient to compute for this case (see \cite{Bag_info} Section IV.C, page 2821).
Future work is planned to compare DLGM and PBN in practice.

\subsection{PBN Deterministic Form}
The deterministic form of the PBN is obtained from the
asymptotic form by replacing $p_n(x;\alpha)$ by their expected values $\lambda_n(\alpha)$.
Interestingly, $\lambda_n(\alpha)$ cancels  $\lambda^{-1}_n(\alpha)$,
leaving $\gamma^{-1}_n(\;)$ as the only non-linearities, except at the visible layer.
This resulting PBN is a deterministic dual to the DAN, which exactly recovers the hidden values.
An arbitrary activation function $\lambda_n(\alpha)$ can be used as long
as $\gamma_n(\bfh_n)$ is defined using the same function. 
Note that $\lambda_n(\alpha)$ must be invertible, so
activations functions like {\it softplus} can be used, but not {\it relu}.

\subsection{Sampling Efficiency}
\label{sampeff}
The sampling efficiency $\epsilon$ is the fraction of times that
the PBN successfully creates a sample of visible data and
depends on the feature generating distribution
$g(\bfz)$ and whether exact (UMS) or deterministic generation is used.
A sampling failure occurs in a UMS block if the set
${\cal M}_m(\bfz_n,e_n)$ has no members, or in the asymptotic or
deterministic PBN if $\gamma^{-1}_{n}({\bf z}_n)$ has no solution.
When sampling fails, it is necessary to re-start
the process by drawing another feature value.
Sampling efficiency, either for UMS or
for deterministic PBN, can be driven towards 1.0 though
training, as will be demonstrated below.

\subsection{PBN Initialization and Training}
\label{jft}
In order to initialze the PBN so it has high sampling efficiency,
the weight matrices should be initialized by principal component analysis (PCA)
of the input data prior to the activation function \footnote{When data is already 
constrained to the range $[0,\; 1]$, as it is
in the MNIST corpus, it is useful to ``gaussianify" the data, mapping to $\mathbb{R}^N$ prior to PCA analysis (See Section \ref{ddesc}).}.
Scaling and bias are then used to provide good ``activation" of $\lambda_n(\;)$.
%
%
In this paper, two types of PBN training are used - deterministic auto-encoder training
and maximum likelihood (ML) training.
In auto-encoder training, the 
DAN is combined with the deterministic PBN to form an auto-encoder (a clockwise circular path at the 
bottom of Figure \ref{pbn_multi}).  Training is accomplished using back-propagation
to minimize total square reconstruction error.
Note that the parameters appear in both PBN and DAN, so the derivative has two terms.
It is critical to have high sampling efficiency for 
auto-encoder training.  In the experiments, $\epsilon$ 
approaches very nearly 1.0 after the first training epoch, 
even for testing data.


In ML training, the log of equation (\ref{cr1a}) is trained for highest average value
by gradient ascent.
%
%
%
We used a special ``uniform assumption" training in which 
the optional activation function (bottom of Figure \ref{pbn_multi}) is applied
to compress the data to the range [0,1],  
and the feature distribution $g(\bfx_3)$ is ignored.
Ignoring the feature distribution is tantamount
to assuming that $g(\bfx_3)=1$, the uniform distribution.
Interestingly, by training this way, a network is produced that,
in fact, produces feature data $\bfx_3$ that is independent uniformly distributed - the 
simplifying assumption becomes fulfilled.

\section{Classification Experiments}
We now compare PBN with a Gaussian mixture model (GMM) in a simple 
classification task.

\subsection{Reduced MNIST Data Description}
\label{ddesc}
For the following experiments, just three characters
``3", ``8", and ``9", of the MNIST handwritten data corpus were used.
Four pixel down-sampling rates were chosen: 1:1, 2:1, 3:1, and 4:1,
resulting in 
data dimensions of 784, 196, 100, and 49.
Since MNIST pixel data is coarsely quantized in the range [0,1],
a dither was applied to the pixel values\footnote{For pixel values
above 0.5, a small exponential-distributed random value was subtracted,
but for pixel values below 0.5, a similar  random value was 
added.}.  To create data in $\mathbb{R}^N$, the inverse sigmoid function was
then applied in order to create ``gaussianified" data with most pixel values in the range -10 and 10.  

\subsection{The 1-layer PBN}
We revisit the 1-layer PBN, which was previously introduced
\cite{BagPBN}. The results of 1-layer PBN experiments 
are relevant to determine if the PBN should be exended to a second layer.
In a multi-layer PBN, a given layer acts as a PDF
model for the features of the up-stream layer. So,
it seems that there is no advantage to adding a layer to a PBN
if a GMM works better than the added layer.
The idea, then is to test a 1-layer PBN against
a GMM as a function of dimension.
This experiment is data-set dependent, so the results
here apply only to MNIST.
%
As a performance benchmark, the GMM was 
applied to the ``gaussianified" data in $\mathbb{R}^N$,
using both diagonal (GMM-D), and full (GMM-F) covariance matrices
\footnote{To avoid singularities, the diagonal elements of the covariance
matrices were multiplied by the factor $(1+\delta)$, where 
$\delta=$ 0.3, 0.3, 0.5, and 0.6 for dimensions 49, 100, 192, and 784, respectively.}.
A separate 1-layer PBN was initialized using PCA,
then trained for each data class 
to maximize the mean log-likelihood using gradient ascent
with ``ADAM" optimization and  L2 regularization using ``uniform assumption" 
training (Section \ref{jft}).  After training, the final activation function was removed,
then $g(\bfz_1)$ was modeled as a GMM.
%
For $N=49, 100, 196, 784$, the number of hidden units
(columns of matrix ${\bf W}$) were 12, 16, 30, and 34, respectively.

Results of the experiment are shown in Figure \ref{pbn_N}.
The PCA-initialized PBN, with no further training
are reported as ``PBN-P", and with training as ``PBN-G".
When comparing ``PBN-P" with ``PBN-G", we can conclude
that ML training greatly improves a PBN.  This means that 
the PDF model offered by a 1-layer PBN is more than 
a just a re-packaged type of Gaussian model or PCA.
The next observation is that the PBN performs better than GMM-F above $N=100$.
%
%
\begin{figure}[h]
  \begin{center}
    \includegraphics[width=3.4in,height=2.4in]{pbn_N.eps}
  \caption{Model comparison as a function of data dimension.} 
  \label{pbn_N}
  \end{center}
\end{figure}
%
%
Both  GMM-F and PBN can model pixel correlation, GMM-F explicitly
using the covariance matrices, and PBN implicitly 
by decorrelating the features, as was noted at the end of Section \ref{jft}.
But, PBN requires $MN$ parameters, versus  the $MN^2$ parameters required for the GMM.
This may explain the advantage of PBN above $N=100$.  
The average sampling efficiency for PBN-G 
was 0.72, 0.85, 0.77, and 0.52 for $N=49,100,192,784$,
respectively.  The worst case change in per-pixel log-likelihood,
is 0.007, so sampling efficiency in Figure \ref{pbn_N} can be essentially
ignored.

\subsection{Multi-layer PBN}
The 1-layer PBNs for $N=196$ and $784$
were extended to a second layer with $16$ and $18$ hidden units, respectively.
The 2-layer PBNs were then trained with an assumption of uniform distribution for $g(\bfx_3)$, then the final activation function was removed and
GMM was used to model the final feature PDF $g(\bfz_2)$.  
Sampling efficiencies were 0.55 and 0.70, respectively,
also negligible.  Performance is shown in Figure \ref{pbn_N} as ``PBG-2-G"
and shows worse performance with respect to 1-layer PBN-G.
This could have been predicted based on Figure
\ref{pbn_N} because the feature dimension is much less than $100$.
Extending the PBN to a second layer would only be effective if the
first layer feature dimension is much larger. 
%

\section{Auto-Encoder (A-E) Experiments}
In the next experiment, a multi-layer
deterministic PBN together with the DAN
are used as an A-E and compared with a standard A-E network 
of the same structure. 
The full $28\times 28$ ($N=784$) data was used.
Separate A-Es were trained on each data 
class to minimize total square error by back-propagation.
ADAM optimization and L2 regularization was used for both network types.
TED (T), sigmoid (S) and softplus (P) activation functions were tried.
The average squared error was measured for testing and training data
and is listed in Table \ref{tab1a}. Although the 
conventional A-E attained a lower squared error
on the training data, it fared much worse on the test data. In contrast,
the PBN had similar squared error on both sets, significantly
out-performing the standard A-E - which
can probably be attributed to (a) that fact that the 
PBN uses the same weights for reconstruction and analysis, and thereby
implements the same task with half the parameters, and (b)
the reconstruction (PBN) is the perfect complement to the 
analysis network (DAN).
Using L2-regularization for conventional A-E did not change this.
The A-E performance for TED and sigmoid was similar, but training took longer for TED.
Sampling efficiency for PBN was 100 percent (no samples that failed reconstruction)
for training, and about 99.9\% (typically 1 sample or less failed) on the
test data.  
\begin{table}
\begin{center}
 \begin{tabular}{|l|l|l|l|l|l|l|}
\hline
Nodes & Act & Type & E-Train & E-Test & Class  \\
\hline
\hline
32-12 & T & A-E  & 7.40 & 10.39 & 1.94\% \\
\hline
32-12  & S & A-E  & 6.73 & 10.79 & 2.97\% \\
\hline
32-12 & T & {\bf PBN}  & 8.63 & {\bf 9.04} & {\bf 1.27\%}  \\
\hline
\hline
36-16 & T & A-E  & 5.84 & 8.21 & 2.57\% \\
\hline
36-16  & S & A-E  & 5.26 & 8.26 & 2.51\% \\
\hline
36-16 & T & {\bf PBN}  & 6.96 & {\bf 7.40} & {\bf 1.70\%}  \\
\hline
\hline
32-16-9 & P & A-E  & 8.27 & 15.3 & 4.4\%  \\
\hline
32-16-9 & P & {\bf PBN}  & 9.95 & {\bf 11.25} & {\bf 0.90\%}  \\
\hline
\end{tabular}
\end{center}
\caption{Total square error for auto-encoder task. Activation functions
(Act) are TED (T), sigmoid (S) and softplus (P)}.
\label{tab1a}
\end{table}
The good generalization of the PBN A-E suggests
using it as a classifier based on minimum reconstruction
error, which we tried.  The results are shown in Table \ref{tab1a} 
in column ``Class".  PBN performed significantly better than A-E,
attaining a very respectable 0.9\%, which handily out-performs 
the standard PBNs in Figure \ref{pbn_N} (denoted by ``PBN A-E").

The deterministic PBN is also useful to generate entirely synthetic data,
In Figure \ref{aenc_syn_pbn}, examples were generated by training a 
GMM on the features (i.e. output of the DAN), then 
passing synthetic features through the PBN. 
The configuration ``32-16-9" with softplus activation was used.  The synthetic samples are sorted in order of decreasing likelihood
(starting from top left), demonstrating the a benefit of 
a tractable likelihood function.
The quality of these samples suggests using the deterministic PBN
in a generative adversarial network (GAN)  - but differing from
a standard GAN in the posession of a tractable LF.
\begin{figure}[h]
  \begin{center}
    \includegraphics[width=3.5in]{aenc_syn_pbn.eps}
  \caption{Data synthesized from determinisic PBN and sorted in order of decreasing
likelihood value.}
  \label{aenc_syn_pbn}
  \end{center}
\end{figure}

\ifdohybrid
\section{Hybrid PBN classifier}
The goal in this experiment is to combine the
results of the last 2 sections by forming a hybrid PBN classifier 
from the 1-layer PBN classifier and the PBN auto-encoder.

For the PBN portion of the hybrid, we used an annealed kernel mixture.
The idea of a PBN kernel mixture is that the features extracted by
a PBN trained on one class might have useful information
regarding another class - especially for poorly formed 
handwritten characters.
The class-specific feature 
mixture (CSFM) \cite{BagIWCCSP,BagAESModelMix,BagUMS}  
approximates the PDF of one class using a mixture of 
all the PBNs, increasing the information
available without increasing the feature dimension.
Annealing improves the linear mixing of the kernels
\cite{BagAESModelMix,BagUMS}.  
We form an annealed kernel mixture of the PBN PDFs (\ref{cr1a}) 
as follows:
\beq
f_m(\bfx; a) = \left(
\sum_{l=1}^c w_{l,m} \; p_p(\bfx; T_l,\smallmath{\hat{p}(\bfz_l|H_m)})^{1/a}
\right)^a,
\label{csfma}
\eeq
where $c$ is the number of classes ($c=3$ here), 
$\bfz_l = T_l(\bfx)$ represents the DAN trained on class $l$,
$\hat{p}(\bfz_l|H_m)$ is the feature PDF estimate for feature $\bfz_l$ and data class $m$,
and $a$ is a heurisic annealing parameter.
The weights are estimated using training data using,
$$\hat{w}_{l,m} = \frac{ \sum_i \; p_p(\bfx_i; T_l,\smallmath{\hat{p}(\bfz_l|H_m)})^{1/a}}
{\sum_i \; \sum_k \; p_p(\bfx_i; T_k,\smallmath{\hat{p}(\bfz_k|H_m)})^{1/a}}.$$
For the deterministic PBN auto-encoder portion of the hybrid, 
we used $h_m(\bfx;b) = {e^{-\|\bfx-\hat{\bfx}_m\|^2/b} \over
\sum_l e^{-\|\bfx-\hat{\bfx}_l\|^2/b}},$
where $\|\bfx-\hat{\bfx}_l\|^2$ is the square auto-encoding error
using auto-encoder trained on class $l$.
The complete hybrid classifier distribution is given by
$p(\bfx|H_m; a,b)=\frac{f_m(\bfx; a) \; h_m(\bfx;b)}{K_m(a,b)}$, where
$K_m(a,b)$ is the normalization constant 
that can be estimated using Monte Carlo integration (MCI)
with the un-annealed mixture $f_m(\bfx; a=1)$ 
acting as proposal distribution\footnote{ 
As a motivation for PBN, we noted that having a
tractable LF avoids the need for MCI, yet here we are using MCI. 
This is not a contradiction.  As dimension increases, the proposal distribution needs 
to be increasingly well matched to the function to be integrated.
To normalize a high-dimensional distribution outright,
for example using GMM as proposal distribution would fail.
But, MCI is useful to normalize 
high-dimensional distributions with tractable LF 
that have been slightly modified,  where the un-modified
distribution acts as proposal distribution.
} \cite{BagUMS}.

Prior to estimating $K_m(a,b)$, classification error was optimized over $a$ (Figure \ref{comb_b} left), without the term  $h_m(\bfx;b)$.
Then, using the value $a=1500$, optimized over $b$
(Figure \ref{comb_b} right), with a resulting minimum error of 0.77\%,
entered in Figure \ref{pbn_N} (left) as ``PBN-H".
With these values of $a$ and $b$, the normalization constant
$K_m(a,b)$ was estimated using Monte Carlo integration
and the normalized LF entered in Figure \ref{pbn_N} (right).
\begin{figure}[h]
  \begin{center}
    \includegraphics[height=1.9in,width=1.1in]{comb_a.eps}
    \includegraphics[height=1.9in,width=1.1in]{comb_b.eps}
    \includegraphics[height=1.9in,width=1.1in]{comb_b_10.eps}
  \caption{Classification error on reduced MNIST
as a function of  $a$ (left) and  $b$ (center), and on full MNIST
as a function of $b$ (right).}
  \label{comb_b}
  \end{center}
\end{figure}

As a final experiment, the hybrid classifier was tried on the full 10-character MNIST data set
to verify the above results and so that it could be compared with published work.
The classification error as a function of $b$ is shown in Figure \ref{comb_b}, right side, and attains a minumum error of 1.25\%
at the same value of $b$, which is comparable to state of the art
fully-connected (non-convolutional) discriminative classifiers not employing pre-processing,
image distortions or deskewing \cite{MNISTResults}.
\fi

\section{Conclusions}
In this paper, a multi-layer PBN has been described,
in its standard, asymptotic, and deterministic forms.
Experiments comparing a 1-layer PBN with a GMM
on a reduced subset of MNIST show that PBN out-performs GMM 
only above a dimension of about 100, which would
suggest using a 2-layer PBN when the output dimension of the first layer is large.
This paper also described a deterministic multi-layer PBN for the
first time and it has been experimentally found to be superior to a standard auto-encoder
when generalizing to test data both in terms of
reconstruction error and classifier performance.
\bibliographystyle{ieeetr}
\bibliography{ppt}
\end{document}

\documentclass[a4paper]{article}

\usepackage{ITG2021_LatexTemplate/itgspeech2021}    
\usepackage{times}            
\usepackage[english]{babel}   
\usepackage[ansinew]{inputenc}
\usepackage[T1]{fontenc}      
\usepackage[sort&compress,numbers]{natbib}      
\usepackage{amsmath,amssymb}
\usepackage{graphicx}
\usepackage[colorlinks=false,pdfborder={0 0 0}]{hyperref}
\usepackage{units}
\usepackage{amsfonts}

\title{New Restricted Boltzmann Machines and Deep Belief Networks for Audio Classification}

\author{Paul M Baggenstoss}

\address{Fraunhofer FKIE, Fraunhoferstr 20,  53343 Wachtberg, Germany \\
Email: p.m.baggenstoss@ieee.org}

\begin{document}

\newtheorem{identity}{Identity}
\newtheorem{hypothesis}{Hypothesis}
\newcommand{\mathtiny}[1]{\mbox{\tiny$#1$}}

\maketitle

\begin{abstract}
In this paper, the deep belief network (DBN), made popular by Hinton in 2006, is re-vitalized using maximum entropy sampling distributions, their corresponding activation functions,  and a new direct training approach based on classifier performance. 
It is shown in keyword classification
experiments that the DBN can compete with state of the art classifiers, and using additive classifier combination, 
improves upon a state of the art deep neural network.
\end{abstract}

\section{Introduction}

\subsection{Background, Motivation, and Prior Work}
The restricted Boltzmann machines (RBM) 
is a widely-used generative stochastic artificial neural network 
that can learn a probability distribution over its set of inputs
\cite{Goodfellow2016}.  The RBM is based on an elegant stochastic model, 
the Gibbs distribution, and is the central idea in a deep belief
network (DBN) made popular by Hinton \cite{HintonDeep06}.
In fact, stacked RBMs and DBNs played a key role in the birth of deep 
learning because they provided a means to pre-train deep networks that 
suffered from vanishing gradients.  At the time, the DBN had comparable performance
to state of the art discriminative classifiers, but has now
been largely forgotten, being made obsolete by 
newer generative and discriminative models.
Deep neural networks (DNNs) using activation functions such as 
rectified linear unit (ReLu), and training
approaches such as batch normalization and drop-out regularization
soon eclipsed the performance of DBNs.  
The original DBN, based on binary Bernoulli
sampling distribution and the sigmoid activation function suffers from
reduced information throughput as a result of suppression of
amplitude information. This especially affects
classifcation of audio signals using spectrograms, which contain subtle amplitude
clues.  

The limitations of the Bernoulli stochastic networks and the
related sigmoid activation function is widely recognized
and continuous-valued alternatives have been suggested - most of these alternatives are
linear on the positive side, similar to 
ReLu and Softplus \cite{NairHintonRelu,Ravanbakhsh,zhou2016softplus}.
Current research lacks a principled approach to 
the selection of stochastic generating distributions (GD)s 
and their associated activation functions.
Currently, continuous-valued GDs are mainly centered around the 
Gaussian distribution \cite{ChoRBM} or derivatives of the Gaussian distribution \cite{Ravanbakhsh}.
The justification for new types of GDs and activation function, remains largely
empirical \cite{YangSoftPlus,WangRelu,NairHintonRelu,jin2016deep,Ravanbakhsh}.

The principle of maximum entropy presents an opportunity for a more principled search
for GDs and activation functions.  The GDs of an RBM are related to the statistical
prediction of the input of the layer, based on the output of the layer
\cite{zhou2016softplus}.  Therefore, we should look closely at how one optimally
predicts the input data based on the network layer output.
A statistical approach to prediction of the input data
requires an {\it a priori}  statistical distribution.  
The principle of maximum entropy
\cite{Jaynes82} proposes to select the distribution  with highest entropy
meeting any known constraints.  
%

\subsection{Proposed Approach}
Given that maximum entropy (MaxEnt) distributions are generally
of the exponential class \cite{Kapur}, it is reasonable to restrict the search to
RBMs based on exponential class of distributions \cite{WellingHinton04}.
By choosing generating distributions and the corresponding
activation functions according the MaxEnt distribution
for a given data range, we arrive at a set of MaxEnt RBMs.

For data in the range [0, 1], we propose the
truncated exponential distribution (TED),
which is the continuous equivalent to the Bernoulli distribution.
The TED RBM has been previously proposed by the author
\cite{BagEusipcoRBM}, but no comparative experiments were conducted. 

For positive-valued data, we propose 
the truncated Gaussian distribution (TG).
The TG distribution has an activation function similar to SoftPlus and
in the limit, to ReLu. Although these distributions have been discussed frequently
\cite{YangSoftPlus,WangRelu,NairHintonRelu,jin2016deep}, 
and the TG distribution and activation function have been
previously described for graphical models \cite{su2016nonlinear},
but as far as we know the TG distribution has not before been proposed for an RBM.

In addition to proposing these new MaxEnt generating
distributions (GDs) for RBMs, we propose a new training approach.
The original training scheme of the DBN involved layer-wise training
of a stacked RBM, followed by fine-tuning the entire network
using a full-network extension of contrastive-divergence (CD) called the ``up-down" algorithm
\cite{HintonDeep06}.  Once trained this way, the classifier results were computed by calculating
the marginalized likelihood function, which is defined up to an unknown constant
and is referred to as the ``free-energy" \cite{HintonDeep06}.
Therefore, the classifier cost was not used at all in training, but computed
only after training was complete. 
With the advent of software frameworks such as Theano and Tensorflow
that symbolically calculate gradients, it is relatively straight-forward to train an entire network
based on any desired cost function. We therefore propose to
fine-tune a DBN based on the final free-energy classifier cost function.


\section{RBMs and MaxEnt GDs}
\label{rbmsec2}
\subsection{Definition of RBM}
\begin{figure}[h!]
  \begin{center}
    \includegraphics[width=3.0in]{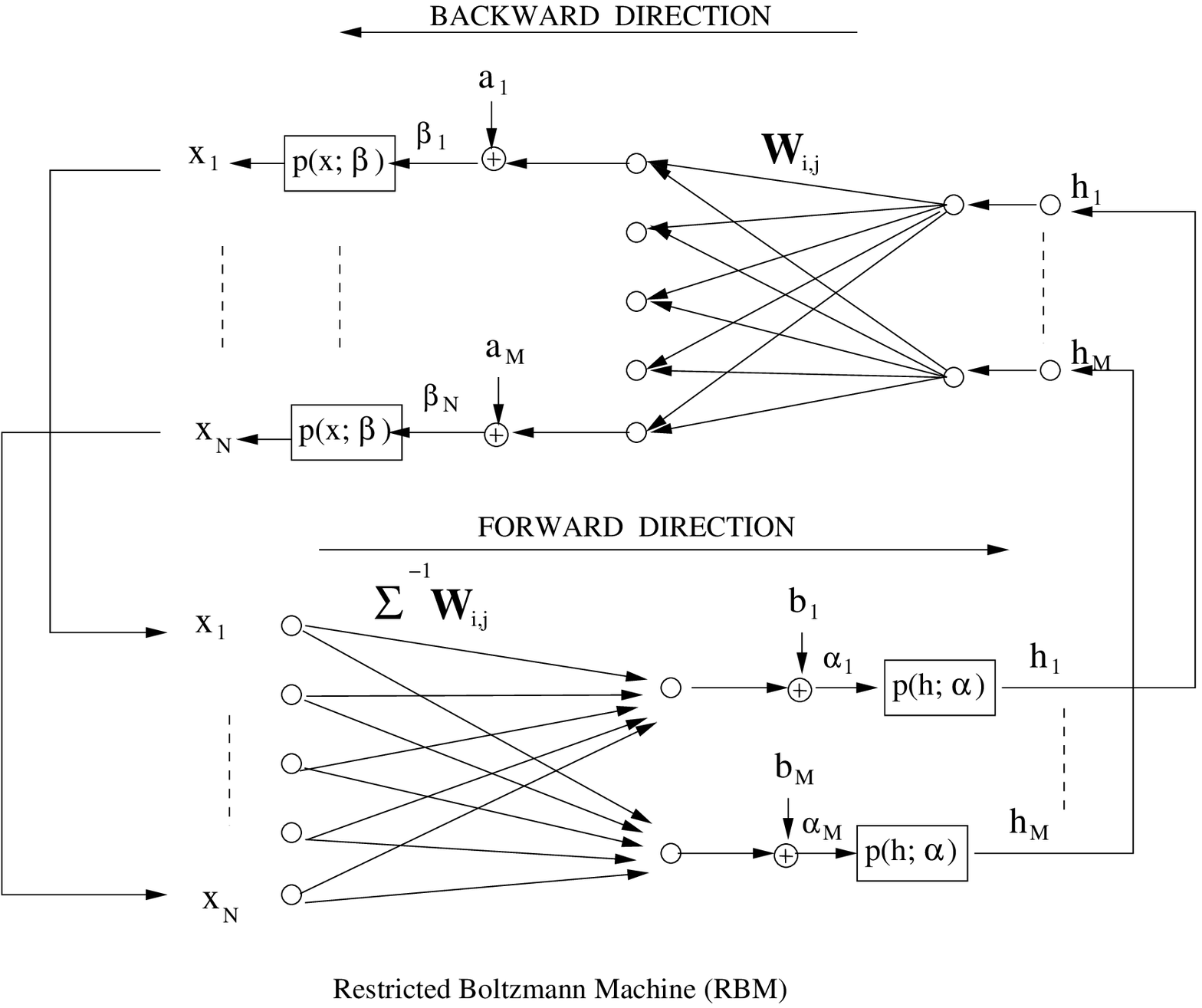}
  \caption{Illustration of an RBM}
  \label{rbm}
  \end{center}
 \vspace{-.1in}
\end{figure}

The RBM estimates a joint distribution between an input (visible) data vector $\bfx\in\mathbb{R}^N$, and
a set of hidden variables  $\bfh\in\mathbb{R}^M$.  
The RBM consists of a pair of stochastic perceptrons, arranged back-to-back, 
and is illustrated in Figure \ref{rbm}.
%
%
In a sampling procedure called Gibbs sampling,
data is created by alternately sampling $\bfx$ and $\bfh$ 
using the conditional distributions $p_h(\bfh|\bfx)$ and
$p_x(\bfx|\bfh)$.  To sample $\bfh$ from the distribution
$p_h(\bfh|\bfx)$, we first multiply $\bfx$ by the 
transpose of the $N \times M$ weight matrix ${\bf W}$, and add a bias vector:
$\balpha = {\bf W}^\prime \bfx + \bfb.$
The variable $\balpha$ is then applied to a generating distribution
(GD) to create the stochastic variable $\bfh$ as $h_i \sim p_h(h|\alpha_i),$  $1\leq i \leq M$.
Note that conditioned on $\bfx$, $\bfh$ is a set of independent random
variables (RV).  To sample $\bfx$ from the distribution
$p_x(\bfx|\bfh)$, we use the analog of the forward sampling process:
$\bbeta = {\bf W} \bfh + \bfa.$ The variable $\bbeta$ is then applied to a generating distribution
$x_j \sim p_x(x|\beta_j),$  $1\leq j \leq N$.  Conditioned on $\bfh$, 
$\bfx$ is a set of independent random variables (RV). 
After many alternating sampling operations, the joint distribution
between $\bfx$ and $\bfh$ converges to the Gibbs distribution
\beq
p(\bfx,\bfh) = \frac{e^{-E(\bfx,\bfh)}}{K},
\label{gibbsd}
\eeq
where the normalizing factor $K$ is generally unknown.
Training an RBM is done using contrastive divergence, which 
is described in detail for exponential-class GDs in \cite{WellingHinton04}.
The algorithm is the same for all GDs.
When training the top-level RBM, it is beneficial to use several
Gibbs iterations \cite{HintonDeep06}.

\subsection{GDs, Activation Functions, and Maximum Entropy}
%
In neural networks, input data or hidden variables in intermediate
layers is often continuous-valued and constrained to a certain data range.
If sigmoid activation is used, data is constrained to the range $[0, \;1]$.
If ReLu or Softplus activations are used, data is constrained to the range $[0, \;\infty]$.
Input data is often unconstrained, so in the range $[-\infty, \;\infty]$.
Each of these data ranges are associated with certain MaxEnt GDs which
are always of the exponential class \cite{Kapur}.
The RBM using exponential-class GDs have been well-defined \cite{WellingHinton04}.
For unconstrained data in the range $[-\infty, \;\infty]$, the Gaussian
distribution has maximum entropy.  The Gaussian RBM is widely used  \cite{ChoRBM}.
For data in the range $[0, \;\infty]$, with constrained mean, there are two options,
with and without constrained variance.  With constrained variance, the MaxEnt distribution
is the truncated Gaussian (TG):
$$p(h;\alpha,\sigma^2)=\frac{\phi(h;\alpha,\sigma^2)}{\Phi(\alpha/\sigma)},$$
where $\phi(h;\alpha,\sigma^2)$ is the Gaussian
$$\phi(h;\alpha,\sigma^2)= (2\pi\sigma^2)^{-1/2} \; e^{-(h-\alpha)^2/(2\sigma^2)},$$
and $\Phi(\;)$ is the cumulative distribution of the standard Gaussian.
For data in the range $[0, \;\infty]$,  without constrained variance,
we have the exponential GD: 
$p(h;\alpha)= \alpha \; e^{-\alpha x},$ however this GD is not as useful as the TG.
For data with constrained mean in the range $[0, \;1]$,
the MaxEnt distribution is the truncated exponential
distribution  (TED) \cite{BagEusipcoRBM,Singh2013}.  The TED GD is 
$p_h(h; \alpha) = C(\alpha) \; e^{\alpha h}$,
where 
$C(\alpha) = \left( \frac{\alpha}{e^{\alpha} - 1}\right)$.

A deterministic variant of the GD is formed by
replacing the stochastic generation with the expected value of the GD
$$\lambda_h(\alpha)=\mathbb{E}(h;\alpha)=\int_h \; h \; p_h(h;\alpha) \; {\rm d} h,$$
which can be seen as an activation function.
The input data activation function $\lambda_x(\beta)$ is similarly defined
replacing $h,\alpha$ with $x,\beta$.
The conditional distributions and activation functions corresponding to each GD are listed in Table \ref{tab1c}.
Interestingly, the TED and TG activations are similar to sigmoid and softplus 
activation functions that are in common use (see Figure \ref{ted-tg}).

\begin{table}[htb]
\begin{center}
 \begin{tabular}{|l|l|l|}
\hline
Type & $p(h;\alpha)$ & Activation Fn. $\lambda(\alpha)$  \\
\hline
Bernoulli & $\sigma((2h-1)\alpha)$ & $\frac{1}{1+e^{-\alpha}}$  \\
\hline
Gauss & $\phi(h;\alpha,\sigma^2)$ & $\alpha$ \\
\hline
TED & $\left( \frac{\alpha}{e^{\alpha} - 1}\right) e^{\alpha h}$  
         & $\frac{e^{\alpha}}{e^{\alpha} - 1}-\frac{1}{\alpha}$  \\ 
\hline
TG & $\frac{\phi(h;\alpha,\sigma^2)}{\Phi(\alpha/\sigma)}$ &  $\alpha + \frac{\phi(\alpha/\sigma;0,1)}{\sigma \; \Phi(\alpha/\sigma)}$ \\
\hline
\end{tabular}
\end{center}
        \caption{Conditional distributions and activation functions by generating distribution (GD).
Elemental distributions are shown. To get the joint conditional distribution, use
$p(\bfh|\bfx)=p(\bfh;\balpha)=\prod_{i=1}^N p(h_i; \alpha_i),$ where $\balpha={\bf W}^\prime \bfx+\bfb.$
}
        \label{tab1c}
\vspace{-.0in}
\end{table}

A deterministic variant of the GD is formed by
replacing the stochastic generation with the expected value of the GD
$$\lambda_h(\alpha)=\mathbb{E}(h;\alpha)=\int_h \; h \; p_h(h;\alpha) \; {\rm d} h,$$
which can be seen as an activation function.
The input data activation function $\lambda_x(\beta)$ is similarly defined
replacing $h,\alpha$ with $x,\beta$.
The activation functions corresponding to each GD are listed in Table \ref{tab1c}.
The TED and TG activations are similar to sigmoid and softplus (see Figure \ref{ted-tg}).
\begin{figure}[h!]
  \begin{center}
    \includegraphics[width=3.1in,height=1.2in]{TED-TG.eps}
  \caption{
eft: TED activation compared to Sigmoid. Right: TG
activation compared to Softplus (reprinted from \cite{BagIcasspPBN}).}
  \label{ted-tg}
  \end{center}
\end{figure}

%

\subsection{Energy Functions}
The Gibbs distribution (\ref{gibbsd}) is based on the energy function,
which in turn depends on which input and output GD are used.
Energy functions have an interaction term  $E_{xh}(\;)$ and two bias terms $E_b(\;)$ as follows:
$$E(\bfx,\bfh)=E_b(\bfx,\bfa,\sigma_x) + E_{xh}(\bfx,\bfh,\sigma_x,\sigma_h) + E_b(\bfh,\bfa,\sigma_h),$$
where
$E_{xh}(\bfx,\bfh,\sigma_x,\sigma_h) = -\frac{ \bfx^\prime {\bf W} \bfh}{\sigma_x \sigma_h}.$
The bias terms depend on the GDs and are given by 
Gaussian and TG: $E_b(\bfx,\bfa,\sigma_x)=\frac{(\bfx-\bfa)^\prime (\bfx-\bfa)}{2 \sigma_x^2},$
and TED: $E_b(\bfx,\bfa,\sigma)= - \frac{\bfx^\prime \bfa}{\sigma}.$
For example, the TG-TED energy function is given by
$$E(\bfx,\bfh)=\frac{(\bfx-\bfa)^\prime (\bfx-\bfa)}{2 \sigma_x^2} -\frac{ \bfx^\prime {\bf W} \bfh}{\sigma_x} - \bfh^\prime \bfb,$$ where we have assumed without loss of generality that $\sigma^2_h=1.$
Compare with Eq. (1) in \cite{ChoRBM}.
\begin{table}[htb]
\begin{center}
 \begin{tabular}{|l|l|l|l|}
\hline
Type & $E_{xh}(\bfx,\bfh)$ & $E_{b}(\bfx,\bfa)$ &  $E_{b}(\bfh,\bfb)$ \\
\hline
Bernoulli & $- \bfx^\prime {\bf W} \bfh$ & $- \bfx^\prime \bfa$ & $- \bfh^\prime \bfb$  \\
\hline
TED & $-\bfx^\prime {\bf W} \bfh$ & $- \bfx^\prime \bfa$ & $- \bfh^\prime \bfb$  \\
\hline
TG & $-\frac{ \bfx^\prime {\bf W} \bfh}{\sigma_x}$ & $\frac{(\bfx-\bfa)^\prime (\bfx-\bfa)}{2 \sigma_x^2}$ & $\frac{(\bfh-\bfb)^\prime (\bfh-\bfb)}{2 }$ \\
\hline
Gauss & $-\frac{ \bfx^\prime {\bf W} \bfh}{\sigma_x }$ & $\frac{(\bfx-\bfa)^\prime (\bfx-\bfa)}{2 \sigma_x^2}$ & $\frac{(\bfh-\bfb)^\prime (\bfh-\bfb)}{2 }$ \\
\hline
\end{tabular}
\end{center}
        \caption{Energy functions by GD. It is assumed that $\sigma^2_h=1$ 
for all GDs and $\sigma^2_x=1$ for the TED GD.}
\vspace{-.2in}
        \label{tab1d}
\end{table}


\section{The Deep Belief Network (DBN)}
A DBN is illustrated in Figure \ref{dbn1}.
The DBN consists of one or more feed-forward perceptron layers,
ending with an RBM at the ``top layer". 
The feed-forward layers are trained as RBMs, then used in their
deterministic form by replacing stochastic generation with deterministic
activation functions.  The top-layer RBM is a classifier formed by appending class labels to the input data
(of the last layer) and training it to estimate the joint Gibbs distribution with the class labels.  
\begin{figure}[h]
\begin{center}
  \includegraphics[width=2.0in,height=2.5in]{dbn2.eps}
\caption{Hinton's DBN.}
\label{dbn1}
\end{center}
\end{figure}
Let $\bfv$ be the
output of the next-to-last layer, and let $\bfy$ be the $M\times 1$ vector of one-hot encoded class labels
(we assume there are $M$ classes).  Then, $\bfy$ is appended
to $\bfv$ to form the input vector to the top-layer RBM $\bfx=[\bfv, \bfy]$.  
Note that the label signal $\bfy$ must be compatible with the
given data range.  The DBN top layer RBM estimates
the joint distribution $p(\bfx,\bfh)$ according to form (\ref{gibbsd}) .
Once trained, (\ref{gibbsd}) is marginalized to get the
distribution of $\bfx$ alone, which is the joint distribution
of $\bfv$ and the class labels $\bfy$: 
\beq
  p(\bfx) =  \frac{1}{K}  {\ds \int_{\bfh}} e^{-E(\bfx,\bfh)} \; {\rm d} \bfh.
  \label{rbmpxHt}
\eeq
This distribution can be solved for in closed form
except for the unknown constant $K$.
To classify, the various assumptions about the class labels $\bfy$ are tested:
$$ \arg \max_m p( [\bfv, \bfy_m] ), $$
where $\bfy_m$ is the one-hot label arrangement for class $m$.
The normalization factor $K$ in (\ref{gibbsd}) is unknown, but
this does not affect the maximization.
The exact form of the marginalized distributions $p(\bfx)$
depend on which GDs are used in the top layer RBM.
The marginals for any combination of input and output GD can be found using Table \ref{tab1a}.
\begin{table}[htb]
\begin{center}
 \begin{tabular}{|l|l|l|}
\hline
Type &  $p_1(\bfx)$ &  $p_2(\bfx)$  \\
\hline
TED &   $ \frac{1}{K}  e^{\bfx^\prime \bfa}$  & $ \frac{1}{C(\balpha)}$ \\
\hline
BERN. &   $\frac{1}{K}  e^{\bfx^\prime\bfa}$  & $\prod_{j=1}^M  \left( 1+e^{\alpha_i} \right)$ \\
 \hline
TG &   $ \frac{1}{K} \; e^{\bfx^\prime \bfa/\sigma_x-\bfx^\prime\bfx/(2\sigma^2_x)}$ & $  (2\pi)^{N/2} \; e^{\balpha^\prime \balpha/2} \; \Phi(\balpha)$ \\
 \hline
Gauss &   $ \frac{1}{K} \; e^{\bfx^\prime \bfa/\sigma_x-\bfx^\prime\bfx/(2\sigma^2_x)}$ & $  (2\pi)^{N/2} \; e^{\balpha^\prime \balpha/2}$ \\
\hline
\end{tabular}
\end{center}
	\caption{Marginal distributions by RBM type. Use $p(\bfx)=p_1(\bfx) p_2(\bfx)$, 
where $p_1(\bfx)$ and $p_2(\bfx)$ are selected according to the input and output GD, respectively. 
Vector function arguments are combined multiplicatively across the elements. The Bernoulli
GD (Bern) is included for completeness.}
	\label{tab1a}
\vspace{-.2in}
\end{table}


\section{Results}
In these experiments, we aim to show the superiority of
the proposed GDs and activation functions with respect to 
combinations of the widely-used Bernoulli/sigmoid and Gaussian/linear.

\subsection{Data}
\label{datsec}
The data was selected to be at the same time realistic, challenging, and relevant to
speech classification.  We selected a subset of the Google speech commands data \cite{Warden2018SpeechCA}.
choosing three pairs of difficult to distinguish words: (``three, tree"),
(``no, go"), and (``bird, bed"), sampled at 16 kHz and segmented into
into 48 ms Hanning-weighted windows shifted by 16 ms.
We used log-MEL band energy features with 20 MEL-spaced
frequency bands and 45 time steps, representing a frequency span of 8 kHz and a time span of 0.72 seconds.
The input dimension was therefore $N=45\times 20=900.$
From each of the six classes, we selected 500 training samples, 150 validation samples, at random.
The remaining samples were used to test, averaging about 1500 per class or about a total of 10000
testing samples.


\subsection{Network}
We used a 5-layer DBN, which was composed of a 4-layer stacked RBM
followed by a top-level RBM with injected one-hot encoded labels.
The first layer had 6 convolutional kernels of size ($13\times 15$) with
convolutional zero-padding, downsampled by ($3\times 4$), producing 6 feature maps of size $15\times 5$.
The second layer had 40 convolutional kernels of size ($7\times 5$) with no convolutional zero-padding, downsampled
by ($2\times 1$), producing 40 feature maps of size $5\times 1$.
The remaining layers of the stacked RBM were dense dense layers of $64$ and $32$ units, respectively. 
The final top-level RBM had 256 stochastic units.
We used truncated Gaussian GD's in the stacked RBM, and the
associated TG activation function (See Table \ref{tab1c}),  but experimented with different
GDs for the hidden units of the top-level RBM.

\subsection{Baseline Classifier}
As a base-line state of the art discriminative classifier, a deep neural network (DNN) was 
trained using the same network structure, but with a last layer of 6
units (for the output cross-entropy classifier layer), using max-pooling instead
of down-sampling, and with dropout regularization.
This classifier achieved an accuracy of 87.5\% on the testing data.
This is good accuracy given the challenging choice of hard to distinguish
word-pairs and existence of many noisy and poorly-segmented samples.

\subsection{DBN Training}
To initialize the DBN, the first four layers were trained layer-wise as 
a stacked-RBM, then fine-tuned using the up-down algorithm \cite{HintonDeep06}.
The classification error performance was then noted for each type of
GD used in the top-level RBM, and is entered in Table \ref{tab1}
in the column  ``DBN". The best performance was seen by TG, follwed by TED, and 
the worst was Bernoulli.
\begin{table}[htb]
\begin{center}
{\footnotesize
 \begin{tabular}{|l|l|l|l|l|l|l|l|l|}
\hline
  & \multicolumn{2}{|c|}{"Bernoulli"} &  \multicolumn{2}{|c|}{"TED"}  & \multicolumn{2}{|c|}{"TG"} & \multicolumn{2}{|c|}{"Gauss"} \\
 \hline
	 & Acc & Rec & Acc & Rec &Acc & Rec &Acc & Rec \\
 \hline
DBN &	 75.4  & 11.0  & 76.9 & 10.0 & 79.1 & 6.0 & 72.0 & 7.1\\
 \hline
FE &	 80.7  & 12.0  & 82.5 & 12.0 & 83.6 & 7.0 & 77.0 & 7.7\\
 \hline
\end{tabular}
}
\end{center}
	\caption{Classification accuracy and mean square reconstruction error 
for DBN and DBN with free-energy training (FE) as a function
of top-layer generating distribution.}
	\label{tab1}
\vspace{-.2in}
\end{table}
\begin{figure}[htb!]
  \begin{center}
    \includegraphics[width=3.0in, height=1.5in]{combXeDbn.eps}
  \caption{Classifier combination results for DNN and DBNFE.}
  \label{combXeDbn}
  \vspace{-.2in}
  \end{center}
\end{figure}

Next, the same network was trained using a combined cost-function. 
Let $p(\bfx_m)$ be the marginal Gibbs distribution of the
top-level RBM taken from Table \ref{tab1a},
where $\bfx_m$ is the input to the top-level RBM, with injected
labels that assume class $m$ is true.  Next, we apply the softmax function
$h_m = \frac{e^{p(\bfx_m)}}{\sum_m e^{p(\bfx_m)}},$
then calculate the usual cross-entropy (XE) cost function based on $\bfh = \{h_m\}$
$C_{xe}={\rm XE}(\bfh,{\bf l}),$ where ${\bf l}$ is the vector of ground-truth one-hot labels.
The network is trained using the usual CD updates 
plus a constant times $\frac{\partial}{\partial \theta} C_{xe}$,
where $\theta$ is a general parameter. Results for this modified DBN training
are entered in Table \ref{tab1} in the column  ``DBNFE". 
Once again, the best performance was seen by TG, follwed by TED, and 
the worst was Bernoulli.  The modified training has resulted in a very significant improvement for all GDs.
Table \ref{tab1} also shows the visible data reconstruction error, which is the
average square reconstruction error obtained when propagating the data deterministically to the top level RBM,
then back to the visible data.  Higher input data reconstruction error is noted for the
modifoed traning.

\subsection{Classifier Combination}
For good results when combining classifiers, two conditions must be met. Fist, the 
classifiers to be combined must have independent information.
Second, they must have comparable performance. Therefore,
having a well-performing generative classifier
is advantageous because it provides a classification
statistic that is largely independent from a discriminative classifier.
In Figure \ref{combXeDbn}, we see the classifier performance as a function
of linear combining factor. As can be observed, a significant performance
improvement can be seen when combining the DBNFE (TG) with the
baseline DNN classifier.


\section{Conclusions}
In this paper, we have proposed to use maximum entropy (MaxEnt) generating distributions (GDs)
for a restricted Boltzmann machine (RBM) to estimate the distribution of
continuous-valued data, then use these new RBMs in a deep belief network (DBN).
In experiments, we have shown the superiority of these new GDs in classifying
spectrograms of spoken words.  In addition, we have proposed a new training approach for DBNs that directly trains the
free-energy classifer cost function.  This method has been shown to be superior
to the usual indirect training approach using contrastive divergence (CD).
\bibliographystyle{ieeetr}
\bibliography{ppt}
\end{document}

continuous data should depend on the
A set of related problems in many fields of current research are 
feature extraction and dimension reduction \cite{Fukunaga1990,BarronMDL},
data compression and reconstruction \cite{PCA,DecoDragan},
and the maximization of information transfer \cite{ZhuRadar,NadalInfo,lin2006conditional}. 
In a broad sense, these problems have the goal of finding some optimal transformation 
or mapping from a high-dimensional input data space, into a lower-dimensional
output space.  The definition of optimality is what separates the approaches.
Although not necessary, we can also add the goal of making the
extracted statistics statistically independent \cite{Rosenblatt,DecoHiOrder,DecoDragan}.
One can argue that principal component analysis (PCA) \cite{PCA} 
solves the problem under all of these optimality criteria
for linear transformations under a simplified Gaussian assumption \cite{DecoDragan}.  
Approaches have been proposed to extend PCA by seeking information-preserving
linear mapping for non-Gaussian assumptions, called independent component analysis 
(ICA) \cite{DecoDragan}, as well as and non-linear transformations \cite{LeeNLDR}.
Except for in the case of invertible transformations \cite{DecoDragan},
no satisfactory solution seems to exist for general
case of non-Gaussian assumptions, non-linear dimension-reducing mappings.
A widely-used criteria for maximizing information transfer
is the InfoMax principle, which seeks to maximize the mutual information (MI) between
the input data and output data \cite{ZhuRadar,NadalInfo,lin2006conditional,DecoDragan}.
However, MI suffers from non-invariance under  1:1 data transformations
\cite{KayInfo}. In the proposed approach, we cast the problem
in terms of PDF estimation, resulting in an approach
invariant to 1:1 transformations and that 
attains optimality with respect to all the desired criteria.

\subsection{Proposed Approach}
We propose an approach based on PDF Projection \cite{BagPDFProj,Bag_info,BagUMS,BagMaxEnt2018},
by posing the problem as one of probability density function
(PDF) estimation. The the distribution of the extracted lower-dimensional statistic is
embedded in a constructed PDF designed to estimate the PDF of the input data.
By estimating the PDF of the input data through
minimzation of the KLD between the true PDF and the constructed PDF 
for a fixed output dimension, information is maximized in such a way that is invariant to
scaling or 1:1 transformations, and independent statistics are extracted. 
The method specializes to PCA for the linear Gaussian case.

\section{Mathematical Foundation}

\subsection{PDF Projection}
Consider an arbitrary deterministic dimension-reducing transformation 
\beq
\bfz = T(\bfx), 
\label{ztr}
\eeq
where $\bfx \in \mathbb{X} \in \mathbb{R}^N$, and $\bfz \in \mathbb{Z} \in \mathbb{R}^M$,  and $M<N$.
Note that  $\mathbb{X}$ and $\mathbb{Z}$ are the supports of variables $\bfx$ and $\bfz$.
Let $g(\bfz)$ be some probability density function (PDF) with support $\mathbb{Z}$.
Then there exists a class of PDFs denoted by ${\cal P}_g$ with support on $\mathbb{X}$
which map to $g(\bfz)$, or in other words, if $p(\bfx) \in {\cal P}_g$, then
it follows that when samples $\bfx$ are drawn from $p(\bfx)$, then
$\bfz=T(\bfx)$ will have exactly distribution $g(\bfz)$.
Because $T(\bfx)$ is dimension-reducing, class ${\cal P}_g$ will have
an infinite number of members.
It can be shown \cite{Bag_info} that any member of ${\cal P}_g$ can be written
\beq
G(\bfx;g,T,H_0) =  \frac{p(\bfx|H_0)}{p(\bfz|H_0)} \; g(\bfz),
\label{ppt0}
\eeq
where $p(\bfx|H_0)$ is some reference distribution with support on
$\mathbb{X}$, and $p(\bfz|H_0)$  is its mapping (through $T(\bfx)$) with support on $\mathbb{Z}$.
A variation of PDF projection, called maximum entropy PDF projection (MEPP)
finds the unique member of ${\cal P}_g$ with maximum entropy \cite{Bag_info}.
To estimate a PDF using the constructed distribution (\ref{ppt0}), we minimize the
``distance" between the true distribution $p(\bfx)$ and the constructed distribution
$G(\bfx;g,T,H_0)$.  To formalize the idea,
we use the usual definition of the Kullback-Leibler divergence (KLD) 
between two PDFs $p(\bfx)$ and $q(\bfx)$ \cite{KayInfo}
\beq
D( p \| q) \defined \mathbb{E}_p \left\{ \log \frac{p(\bfx)}{q(\bfx)}\right\}. 
\label{kld}
\eeq
Therefore, PDF estimation is equivalent to
\beq
\min_{g,T,H_0} \left\{ D\left( p(\bfx) \| G(\bfx;g,T,H_0) \right) \right\}.
\label{trainpdf}
\eeq
Using (\ref{kld}), and recognizing that $\mathbb{E}_p \left\{ \log p(\bfx) \right\}$ is fixed,
PDF estimation is the same as 
\beq
\max_{g,T,H_0} \mathbb{E}_p \left\{ \log G(\bfx;g,T,H_0) \right\},
\label{trainpdf2}
\eeq
which is equivalent to maximum likelihood estimation of the given parameters.

\subsection{Fulfillment of Assumptions Principle}
Let $p_T(\bfz)$ be defined as the  mapping of the true PDF $p(\bfx)$ through transformation $T(\bfx)$.
This is the true distribution of the features $\bfz=T(\bfx)$, given that $\bfx$ is drawn from
$p(\bfx)$ for a given $T(\bfx)$.  The {\it principle of fulfillment of assumptions} states that if an arbitrary distribution
$g(\bfz)$ is assumed for the feature PDF in (\ref{ppt0}), then after training
(\ref{trainpdf}) or (\ref{trainpdf2}), we have
\beq
p_T(\bfz) \rightarrow g(\bfz).
\label{fap}
\eeq
{\it Proof by contradiction.} We first assume that $p_T(\bfz) \neq g(\bfz)$.
Now consider a 1:1 mapping from $\mathbb{R}^M$ to $\mathbb{R}^M$,
denoted by $\bfw = H(\bfz)$, with Jacobian $J_H(\bfz)=\left[\frac{\partial w_i}{\partial z_j}\right]$,
such that $|J_H(\bfz)|=\frac{p_T(\bfz)}{g(\bfw)}.$
It can then be shown that 
$\mathbb{E}_p \left\{ \log G(\bfx;g,T_H,H_0) \right\} > \mathbb{E}_p \left\{ \log G(\bfx;g,T,H_0) \right\}$,
where $T_H$ is the combined transformation $T_H(\bfx) =H(T(\bfx))$, 
contradicting the assumption that (\ref{trainpdf2}) had been maximized. 

The importance of the fulfillment of assumptions principle is that we can fix
$g(\bfz)$ {\bf without loss of generality} to equal any arbitrary distribution,
thereby removing $g(\bfz)$ from the set of parameters that need to be optimized.
We then re-state the problem to be solved as:
\beq
\min_{T,H_0} \left\{ D\left( p(\bfx) \| G(\bfx;g_0,T,H_0) \right) \right\},
\label{trainpdf3}
\eeq
where $g_0(\bfz)$ is a fixed reference distribution, such as independent and identically distributed
({\it iid}) uniform or Gaussian random vatiables (RV).
This has two significant advantages: (a) it makes the proposed approach
invariant to 1:1 transformations of the features, and (b) allows
us to specify {\it iid} output RVs, thereby attaining two stated objectives simultaneously.
The maximization of information is justified next.

\subsection{A New Information Maximization Criterion}
By expanding (\ref{trainpdf3}), we arrive at
\beq
\begin{array}{l}
D\left( p(\bfx) \| G(\bfx;g_0,T,H_0) \right)  =  
  \mathbb{E}_p\left\{ \log \frac{p(\bfx)}{p(\bfx|H_0)}  - \log \frac{g_0(\bfz)}{p(\bfz|H_0)} \right\} \\
 \;\;\;\;\;\;\;=  \mathbb{E}_p\left\{ \log \frac{p(\bfx)}{p(\bfx|H_0)}\right\}-\mathbb{E}_p\left\{ \log \frac{g_0(\bfz)}{p(\bfz|H_0)} \right\} .
\end{array}
\label{pst2}
\eeq
But notice that in the second term, $\mathbb{E}_p\left\{ \;\right\}$ can be taken with respect to either the distribution
of $\bfx$ or $\bfz$, since $\bfz$ is derived from $\bfx$. Therefore,
the problem reduces to the minimization of 
\beq
\begin{array}{l}
D\left( p(\bfx) \| G(\bfx;g_0,T,H_0) \right)  =   \\
 \;\;\;\;=  D\left( p(\bfx) \| p(\bfx|H_0) \right) - D\left( g_0(\bfz) \| p(\bfz|H_0) \right).
\end{array}
\label{pst3}
\eeq

{\it Interpretation}.  The first term is the KLD between the true PDF $p(\bfx)$
and the reference distribution $p(\bfx|H_0)$, what we could call
the ``distinctness" of the pair $p(\bfx)$ and $p(\bfx|H_0)$.
The second term is the same, but seen from the feature space,
and negative.  The second term has a maximum value equal to the first term,
thereby making (\ref{pst3}) a positive quantity, equal to zero
only in the special case that  $G(\bfx;g_0,T,H_0)=p(\bfx)$ [TBD: show this].
The optimization problem requires us to choose a reference distribution that
minimizes the ``distinctness" of $p(\bfx)$ and $p(\bfx|H_0)$
in the input space, while at the same time choosing a transformation 
$T$ that retains the ``distinctness" as seen from the feature space.
This makes intuitive sense because it requires conserving
the information that allows us to distiguish data from the two
distributions $p(\bfx)$ and $p(\bfx|H_0)$.

The choice of $p(\bfx|H_0)$ complicates the optimization.
In PDF projection, $p(\bfx|H_0)$ can be interpreted as a
Bayesian prior [TBD: show this].
Indeed, choosing $p(\bfx|H_0)=p(\bfx)$ results in the 
ideal condition where (\ref{pst3}) is zero [TBD: show this].
This is a classic dilemma that can be avoided by
resorting to the principle of maximum entropy \cite{Jaynes82} to
specify $p(\bfx|H_0)$.  By doing this, 
we can concentrate on the choice of 
transformation $T$.  

For a fixed transformation $T$, choosing $H_0$ creates contradictory goals,
seeking an $H_0$ that is ``close" to $p(\bfx)$ in the input space,
but ``far" from $p_T(\bfx)$ in the feature space.
Is the maximum entropy choice not the best compromise?
So, if MaxEnt is the best choice regardless of $T$, then 
maybe it does make sense to choose a MaxEnt prior $H_0$ and
concentrate on the coice of $T$.  Specifically, we seek a transformation $T$ that
maximizes the ``distinctness" between the true distribution
and reference distribution in the feature space.

[Are there any other interpretations of (\ref{pst3}) ??]

\subsection{Linear Gaussian Case}
We now show that the proposed approach corresponds to principal component analysis (PCA).
Consider an $N\times 1$ vector $\bfx$ and two
Gaussian distributions 
$p(\bfx) \sim N({\bf 0},{\bf R}),$ and 
$p_0(\bfx) \sim N({\bf 0},{\bf I}).$ 
Now consider a dimension-reducing linear transformation
represented by the $N\times M$ matrix ${\bf A}$, where
$M<N$. Let $\bfz={\bf A}^\prime \bfx$. 
The two distributions $p(\bfx)$ and $p_0(\bfx)$ 
are mapped to the featue space as
$p(\bfz) \sim N({\bf 0},{\bf A}^\prime  {\bf R} {\bf A} ),$ and 
$p_0(\bfz) \sim N({\bf 0},{\bf A}^\prime   {\bf A} ).$ 
The KLD between $p(\bfz)$ and $p_0(\bfz)$ is given by
$$D(\;p(\bfz)\;\| \; p_0(\bfz) \;)= 
\mathbb{E}_p\left\{ \log\left( { p(\bfz)\over p_0(\bfz)}\right) \right\},$$
where $\mathbb{E}_p\left\{\;\right\}$ is the expectation taken over
distribution $p(\bfz)$.
The problem is to maximize $D(\;p(\bfz)\;\| \; p_0(\bfz) \;)$ by choice of 
${\bf A}$.

{\bf Solution}
It is easily shown (\cite{KayInfo}, equ. 3.1) that
$$D(\;p(\bfz)\;\| \; p_0(\bfz) \;)= \frac{1}{2} {\rm tr}\left( {\bf A}^\prime  {\bf R} {\bf A} ({\bf A}^\prime   {\bf A})^{-1}\right)
- \frac{1}{2} \log \frac{ {\rm det}\left( {\bf A}^\prime  {\bf R} {\bf A}  \right)}{ {\rm det}\left( {\bf A}^\prime   {\bf A}\right)} - \frac{M}{2}.$$

We first simplify by using the SVD of ${\bf A}$ using
the SVD $ {\bf A}  = {\bf U} {\bf S} {\bf V}^\prime,$
where we use the {\it reduced} SVD, where ${\bf U}$ is dimension
$N\times M$ and ${\bf S}$ is $M\times M$, leaving out the singular vectors with zero singular value.
It can be shown that 
$$ {\bf A}^\prime  {\bf R} {\bf A} ({\bf A}^\prime   {\bf A})^{-1} = 
{\bf V} {\bf S} {\bf U}^\prime {\bf R} {\bf U} {\bf S} {\bf V}^\prime {\bf V} {\bf S}^{-2} {\bf V}^\prime.$$
Using the properties of trace and simplifying,
$$ {\rm tr}\left( {\bf A}^\prime  {\bf R} {\bf A} ({\bf A}^\prime   {\bf A})^{-1}\right) = 
{\rm tr}\left( {\bf U}^\prime {\bf R} {\bf U} \right),$$
and
$$\frac{ {\rm det}\left( {\bf A}^\prime  {\bf R} {\bf A}  \right)}{ {\rm det}\left( {\bf A}^\prime   {\bf A}\right)} =
{\rm det}\left( {\bf U}^\prime  {\bf R} {\bf U} \right).$$
So, we are left with
\beq
D(p(\bfz)\| p_0(\bfz))=  \frac{1}{2} \left[  {\rm tr}\left( {\bf U}^\prime {\bf R} {\bf U} \right)
- \log  {\rm det}\left( {\bf U}^\prime  {\bf R} {\bf U}  \right) - M\right].
\label{kldeq}
\eeq

Both the first and second terms in (\ref{kldeq}) have an extremum
when the columns of ${\bf U}$ are the top $M$ eigenvectors of
matrix ${\bf R}$, which we call the {\it test case}.  
At the test case, the first term in (\ref{kldeq}) 
is the sum of the top $M$ eigenvalues, 
and the second term is the (negative of) the log of the product
of the top $M$ eigenvalues.

Let ${\bf R} = {\bf V}  \bLambda {\bf V}^\prime$,
where $\bLambda$ is the diagonal matrix of eigenvectors $\lambda_1,\lambda_1 \ldots \lambda_M$, 
where we assume $\lambda_i > 1$. This is a reasonable assumption
when using PCA since if any eigenvalues of
${\bf R}$ are less than one, the corresponding eigenvectors 
can become more important than some of the
top eigenvectors for distinguishing covariance ${\bf R}$ from ${\bf I}_N$.
Then the test case is such that ${\bf U}={\bf V}_M$, where
${\bf V}_M$ is the set of top $M$ eigenvectors of ${\bf R}$.
Because of the difference in sign of the two terms, it is not clear
if this condition produces a global maximum of the equation
as a whole.

With the shorthand notation $d \defined \frac{\partial}{\partial t}$ where
$t$ is a scalar parameter of  ${\bf U}$, the derivatives are:
$$ d \;  {\rm tr}\left( {\bf U}^\prime {\bf R} {\bf U} \right)
= 2 {\rm tr}\left( d{\bf U}^\prime {\bf R} {\bf U} \right),$$
$$  d \;  \log {\rm det}\left( {\bf U}^\prime  {\bf R} {\bf U}  \right)
= 2 {\rm tr}\left( ({\bf U}^\prime {\bf R} {\bf U})^{-1}   d{\bf U}^\prime {\bf R} {\bf U} \right).$$

At the test case, ${\bf U}^\prime {\bf R} {\bf U} = \bLambda_M$ and
${\bf R} {\bf U}={\bf V} \bLambda {\bf V}^\prime {\bf V}_M = {\bf V}_M \bLambda_M,$
where $\bLambda_M$ is the $M\times M$ diagonal matrix of the top $M$ eigenvalues,
resulting in 
$$ d \;  {\rm tr}\left( {\bf U}^\prime {\bf R} {\bf U} \right)
= 2 {\rm tr}\left( d{\bf U}^\prime {\bf V}_M  \bLambda_M \right),$$
and 
$$ d \;  \log {\rm det}\left( {\bf U}^\prime  {\bf R} {\bf U}  \right)
= 2 {\rm tr}\left( \bLambda_M^{-1}   d{\bf U}^\prime {\bf V}_M  \bLambda_M \right) =  2 {\rm tr}\left( d{\bf U}^\prime {\bf V}_M  \right).$$

Because the first term in (\ref{kldeq}) has an extremum at the test case,
 ${\rm tr}\left( d{\bf U}^\prime {\bf V}_M  \bLambda_M \right)=0$ ,
from which we can conclude that $ d{\bf U}^\prime {\bf V}_M$ has a zero diagonal.
Then, ${\rm tr}\left( d{\bf U}^\prime {\bf V}_M  \right)=0$, validating our
assertion that the second term in (\ref{kldeq}) also has an extremum at the test case.
To see if (\ref{kldeq}) as a whole has a maximum or minimum at the test case, we need the second derivatives.
With the shorthand notation $dd \defined \frac{\partial^2}{\partial t \partial u} $ where
$t,u$ are parameters of ${\bf U}$, the second derivatives are:
$$dd \; {\rm tr}\left( {\bf U}^\prime {\bf R} {\bf U} \right)
= 2 {\rm tr}\left( d{\bf U}^\prime {\bf R}  d{\bf U} \right) + 2 {\rm tr}\left( dd{\bf U}^\prime {\bf R} {\bf U}  \right),$$
and
$$ dd \; \log {\rm det}\left( {\bf U}^\prime  {\bf R} {\bf U}  \right)
= $$
$$ \;\;\; -4 {\rm tr}\left( ({\bf U}^\prime {\bf R} {\bf U})^{-1}   (d{\bf U}^\prime {\bf R} {\bf U})
({\bf U}^\prime {\bf R} {\bf U})^{-1}  (d{\bf U}^\prime {\bf R} {\bf U})  \right)$$
$$ \;\;\;+ 2 {\rm tr}\left( ({\bf U}^\prime {\bf R} {\bf U})^{-1}   (dd{\bf U}^\prime {\bf R} {\bf U}+d{\bf U}^\prime {\bf R} d{\bf U}) \right).$$

Under test case,  
$$dd \;  {\rm tr}\left( {\bf U}^\prime {\bf R} {\bf U} \right)
= 2 {\rm tr}\left( d{\bf U}^\prime {\bf V}_M \bLambda_M {\bf V}_M^\prime  d{\bf U} \right) 
+ 2 {\rm tr}\left( dd{\bf U}^\prime {\bf V}_M \bLambda_M  \right),$$
and
$$dd \; \log {\rm det}\left( {\bf U}^\prime  {\bf R} {\bf U}  \right)
= $$
$$ \;\;\; -4 {\rm tr}\left( \bLambda_M^{-1}  (d{\bf U}^\prime {\bf V}_M  \bLambda_M)
\bLambda_M^{-1}  (d{\bf U}^\prime {\bf V}_M \bLambda_M)  \right)$$
$$ \;\;\;+ 2 {\rm tr}\left( \bLambda_M^{-1}  (dd{\bf U}^\prime {\bf V}_M \bLambda_M +d{\bf U}^\prime  {\bf V}_M \bLambda_M {\bf V}_M^\prime  d{\bf U}) \right).$$

Using that $ d{\bf U}^\prime {\bf V}_M$ has a zero diagonal, and some magic,
we get
$$\frac{\partial^2}{\partial t \partial u}  {\rm tr}\left( {\bf U}^\prime {\bf R} {\bf U} \right)
=  2 {\rm tr}\left( dd{\bf U}^\prime {\bf V}_M \bLambda_M  \right),$$
and
$$\frac{\partial^2}{\partial t \partial u} \log {\rm det}\left( {\bf U}^\prime  {\bf R} {\bf U}  \right)
=  2 {\rm tr}\left( \bLambda_M^{-1}  (dd{\bf U}^\prime {\bf V}_M \bLambda_M) \right)=  4 {\rm tr}\left( dd{\bf U}^\prime {\bf V}_M \right).$$
We then have, at the test case,
$$\frac{\partial^2}{\partial t \partial u} D(p(\bfz)\| p_0(\bfz))=  
 {\rm tr}\left( dd{\bf U}^\prime {\bf V}_M \bLambda_M  \right)
- 2 {\rm tr}\left( dd{\bf U}^\prime {\bf V}_M \right)$$
$$ \; \; \; \; = {\rm tr}\left( (dd{\bf U}^\prime {\bf V}_M) (\bLambda_M-2{\bf I}_M)\right).$$
Since we have assumed that the term ${\rm tr}\left( {\bf U}^\prime {\bf R} {\bf U} \right)$
has a local maximum, its second derivative must be negtive,
$dd \; {\rm tr}\left( {\bf U}^\prime {\bf R} {\bf U} \right) < 0$,
we will have a local maximum for formula (\ref{kldeq})
as long as the eigenvalues are all greater than 2 [ should this be 1?].

\section{Examples}

\bibliographystyle{ieeetr}
\bibliography{ppt}
\end{document}

statistical and
Information maximization
Much has been published on the comparison of generative and discriminative classifiers.
The widespread view is that discriminative classifiers
generalize better when sufficient labeled training data is available \cite{Lasserre06}.
Despite their success,  it has been recognized that
discriminative methods have flaws, vividly demonstrated by
{\it adversarial sampling} \cite{MayerAdvSamp}, a technique in which
small, almost imperceptible changes to the input data cause
false classifications.  Because generative classifiers are based on a model of the
underlying data distribution, they are immune to adversarial sampling
and can complement discriminative classifiers.
As a result, there are a large number of methods that seek to combine generative and discriminative
classifiers \cite{jaakkola98exploiting,raina03classification,fng01,Fujino05,Holub08,Bosch08,Lasserre06},
or to combine discriminative and generative training \cite{Lasserre06,Minka05,BishopGenDisc}
The weakness of generative classifiers stems from the need to estimate the data distribution,
a very difficult task that is unecessary when just classifying between known
data classes \cite{Vapnik99}.  Deep layered generative networks
are a step in the right direction because they
can model complex data generation processes, but they
have a serious flaw: the data distribution,
also called likelihood function (LF) is intractible because
the hidden variables are jointly distributed with the input data and must be integrated out.
Such networks need to be trained using surrogate cost functions such as contrastive divergence 
to train restricted Boltzmann machines \cite{WellingHinton04,HintonDeep06}, and Kullback
Leibler divergence to train variational auto-encoder (VAE) \cite{pmlr-v32-rezende14},
or an adversarial discriminative network to train generative adversarial networks 
(GAN)  \cite{GoodfellowGAN2014}.

In summary, there is a need for better generative models
with tractable LF  that can be combined with discriminative approaches.
The newly introduced layered generative network called projected belief network (PBN) 
stands out as a potentially better choice to achieve these goals.
The PBN is a deep layered generative network (DLGN), so can model complex generative processes,
but differs from all other DLGNs in three ways. First, it has a tractable likelihood function, 
so can be trained directly. Using the tractable LF, it can detect out-of-set samples 
(outliers that are outside of the set of training classes).
Second, the PBN is based on a feed-forward neural network (FF-NN), so it can share 
an embodiment with a discriminative classifier (i.e. it is a single network
that is both a complete generative model and a discriminative classifier).
Third, the discribution of the output layer (output variables of the last layer) 
is embedded as a separate factor in the likelihood function, so
can be used to inject discriminative behavior into the network
without compromising the generative model.
For these three reason, the PBN is a more direct way to introduce the
advantages of generative models into a discriminative classifier, or vice-versa.

\subsection{Main Idea}
The goal of this paper is to construct a layered generative network with tractable
likelihood function that is at the same time a fully discriminative classifier.
As we stated, the PBN is based on a feed-forward neural network (FF-NN).
Figure \ref{asy0} shows a simple 3-layer feed-forward neural network (FF-NN).
\begin{figure}[h!]
  \begin{center}
    \includegraphics[width=3.5in]{asy0.eps}
  \caption{A feed-forward neural network (FF-NN).  This FF-NN can be a discriminative classifier
if $\lambda_4$ is the {\it softmax} function and the output
box is the cross-entropy cost function.  It can also be a generative model if viewed as  PBN
 and  the output box is the output prior distribution  $g(\bfx_4)$.}
  \label{asy0}
  \end{center}
\end{figure}
Each layer $l$ consists of a linear transformation (represented by matrix ${\bf W}_l$),
a bias $\bfb_l$ and an activation function $\lambda_{l+1}(\;)$.
The linear transformation can be fully-connected or convolutional,
but must have total output dimension lower than the input dimension.
The output layer is required to have a total dimension equal to the number of classes.
Therefore, if the final activation function were {\bf softmax}, then this 
network could be trained as a traditional classifier using cross-entropy 
cost function.  On the other hand, it can also be viewed as a projected belief network (PBN) \cite{BagPBN,BagEusipcoPBN}.
As such, it has the likelihood function (see \cite{BagEusipcoPBN}) given by
\beq
\begin{array}{l}
p_p(\bfx_1; T, g) = 
\frac{1}{\ds \epsilon} \; \frac{\ds p(\bfx_1 ; H_{0,1})}{\ds p(\bfz_1 ; H_{0,1})} \;  |{\bf J}_{\bfz_1 \bfx_2}| \;  \cdot \\  
  \;\;\;\;\;\;\;\;\; \frac{\ds p(\bfx_2 ; H_{0,2})}{\ds p(\bfz_2 ; H_{0,2})} \; |{\bf J}_{\bfz_2 \bfx_3}| \;
	\frac{\ds p(\bfx_3 ; H_{0,3})}{\ds p(\bfz_3 ; H_{0,3})} \;  |{\bf J}_{\bfz_3 \bfx_4}| \; g(\bfx_4),
\end{array}
\label{cr1a}
\eeq
where $\epsilon$ is the {\it sampling efficiency} \cite{BagEusipcoPBN}
that we can assume to be 1.0, $p(\bfx_l ; H_{0,l})$ is the assumed 
prior distribution for the input to layer $l$, denoted by  $\bfx_l$,
 $p(\bfz_l ; H_{0,l})$ is the distribution of $\bfz_l$ under the assumption that
$\bfx_l$ is distributed according to $p(\bfx_l ; H_{0,l})$, and where
$g(\bfx_{L+1})$ is the assumed prior for the output of a network.
The PBN is trained by maximizing the mean of the log of (\ref{cr1a})
using stochastic gradient ascent.  To be a PBN, however, the network must have decreasing 
dimension in each layer.
In the following, we will describe how to create a prior $g(\bfx_{L+1})$ 
so that the same network can be viewed as a discriminative
classifier and as a generative PBN. In other words, the prior $g(\bfx_{L+1})$ will assume the role of the cross-entropy cost function,
so will result in a generative/discriminative network.

\section{Technical Approach}
\subsection{Output Non-Linearity and Prior}
In order to create a PBN that is compatible with a discriminative classifier,
we used the truncated exponential distribution (TED)  activation
function (non-linearity) \cite{BagEusipcoPBN,BagUMS}
given by
$\lambda(\alpha) =  \frac{e^{\alpha}}{e^{\alpha} - 1}-\frac{1}{\alpha}$,
which is similar to sigmoid, producing output in [0, 1]. We 
used a TED prior output distribution $g(\bfx_{L+1})$  given by
$g( \bfx ; \balpha) = 
\prod_i \; \left(
\frac{\alpha_i}{e^{\alpha_i} - 1}\right)  \; e^{\alpha_i x_i}, 
$
where $\balpha$ depends on the class labels (the ground-truth label for input
data $\bfx_1$).  The relationship is $\alpha_i = 2 C (l_i-.5)$,
where $[l_1, l_2 \ldots l_M]$ are the one-hot label encodings,
so $\alpha_i$ has values $C$ or $-C$..
Recall the output dimension $M$ is also the number of classes.
An approximation to this prior for large $C$ is to add
a value of $log(C)$ to the LF when output $x_i$ matches $l_i$, and a value
of $log(C)-C$ if it is the logical inverse of $l_i$. 
Training the network to maximize the average of the log of (\ref{cr1a}) 
can be interpreted as discriminative (through term $g(\bfx_{L+1})$) and generative
through the remaining terms.  
The degree of discriminative training can be varied
by changing $C$.


\subsection{MaxEnt Reconstruction and Synthesis}
\label{reconsec}
We now investigate a distinctly generative property of the
PBN : visible data reconstruction from hidden variables.
Input data can be randomly synthesized or
reconstructed from the output of any layer of the FF-NN.

Unlike other generative networks, the PBN is not an
explicit generative network, it operates implicitly 
by ``backing up" through a FF-NN.
In each layer, the FF-NN operates on the input $\bfx$ by
dimension-reducing linear transformation $\bfz={\bf W}^\prime \bfx$.
To ``back up", it is necessary to determine the set ${\cal M}(\bfz)$
of possible input samples $\bfx$  that ``could have" produced
$\bfz$. In other words, $${\cal M}(\bfz) = \{ \bfx :
{\bf W}^\prime \bfz=\bfx\}.$$  A sample is selected from ${\cal M}(\bfz)$
with probability density proportional to the prior distribution $p_0(\bfx)$. 
This is called ``MaxEnt" sampling because $p_0(\bfx)$ is the maximum entropy
prior given the range of $\bfx$.  It is also called uniform manifold sampling (UMS) because under certain
conditions, $p_0(\bfx)$ has constant value on ${\cal M}(\bfz)$.
Sampling requires a type of Markov chain Monte-Carlo (MCMC) \cite{BagUMS}.
For deterministic reconstruction, we select $\bfx$ to be the
centroid of ${\cal M}(\bfz)$, which is also the conditional mean
  $\hat{\bfx}|\bfz = {\mathbb E}(\bfx|\bfz).$

  This conditional mean can be found in closed form for
  a range of MaxEnt priors \cite{BagIcasspPBN,BagEusipcoPBN,BagUMS}. 
  It is given by $\hat{\bfx}|\bfz =  \lambda\left({\bf W} \hat{\bfh} \right),$
where $\hat{\bfh}$ is the solution of the equation 
\beq
{\bf W}^\prime \lambda\left({\bf W} \bfh \right)=\bfz.
\label{hsola}
\eeq
This solution is guaranteed to exist as long as $\bfx$ is in the support $p_0(\bfx)$
and is also the saddle-point for the saddle-point approximation to $p_0(\bfz)$  \cite{BagIcasspPBN}.
For the simplest case of Gaussian MaxEnt prior, the activation function is linear, $\lambda(\alpha)=\alpha$,
and the reconstruction is just least-squares, $\hat{\bfx}|\bfz = {\bf W} \left({\bf W}^\prime {\bf W}\right)^{-1}\bfz.$
For positive-valued data, we use the truncated Gaussian prior \cite{BagIcasspPBN}, and for data in
$[0,\; 1]$, we use the uniform prior \cite{BagIcasspPBN}. 

Starting at any layer output, one can proceed in the backward direction
up the network, always increasing the dimension, until the visible data
is reconstructed.  Note that $\hat{\bfh}$ is only guaranteed to exist if 
$\bfz={\bf W}^\prime \bfx$ for some $\bfx$ in the support $p_0(\bfx)$.
But, when reconstructing from more than one layer, this requirement is not always
met, so the reconstruction chain could fail (see {\it sampling efficiency} in \cite{BagEusipcoPBN}).  
However, after the network is trained, reconstruction failure is rare \cite{BagEusipcoPBN},
and often means the input sample is mal-formed.

Note that there are two possible methods to reconstruct $\bfx$ from $\bfz$, (a) 
random sampling in ${\cal M}(\bfz)$ by MCMC , and (b) deterministically
selecting the centroid   $\hat{\bfx}|\bfz$.
In the following, we use the approach of random sampling the last 2 layers,
then deterministic reconstruction back to the visible data.

\subsection{PBN Properties}
The proposed method differs significantly with other methods
of combining the roles of generative and discriminative networks
that are available in the field because the
discriminative influence is added into the output prior
and does not disturb the ``purity" of the generative network.
In other words, there is no compromise between generative and
discriminative training or structure, they are both 
contained in one network and one cost function!

When reconstructing visible data from hidden variables, 
the synthesized data, when applied to the feed-forward
network, produces exactly the same hidden variables. This property of hidden variable recovery
is interesting and is true of no other layered generative network.

When the discriminative cost function is ``satisfied" (the training data is almost 
completely separated), then the generative cost dominates, so the network becomes the best possible
PBN that at the same tiome separates the data.
This can be seen as a generative regularization effect.

\section{Classification of Spectrograms of Words Commands}
\subsection{Data set}
The data was selected to be at the same time relevant, realistic,
and challenging.  We selected a subset of the Google speech commands data \cite{GoogleKW},
choosing three pairs of difficult to distinguish words: ``three, tree",
``no, go", and ``bird, bed", sampled at 16 kHz and segmented into 
into 48 ms Hanning-weighted windows shifted by 16 ms.  
We used log-MEL band energy features with 20 MEL-spaced
frequency bands and 45 time steps, representing a frequency span of 8 kHz and a time span of 0.72 seconds.
The input dimension was therefore $N=45\times 20=900.$
From each of the six classes, we selected 500 training samples, 150 validation samples, at random.
The remaining samples were used to test, averaging about 1500 per class or about a total of 10000
testing samples.

\subsection{Network}
A separate network was trained on each word pair.
The networks had $L=5$ layers.  The first layer was convolutional with
36 $21\times 16$ convolutional kernels using ``valid" border mode and
$6\times 4$ downsampling (not pooled, just down-sampled), thus producing
36 $5\times 2$ output feature maps, or a total output dimension of 360.
The remaining 3 layers were fully-connected with 100, 32, and 16 hidden neurons.
The output layer had 2 neurons, matching the number of classes. 
Note that we sought to reduce the dimension in each layer by at least a factor of 2.
The layer output activation functions were linear, TG, TG, TG, and softmax,
where TG is the truncated Gaussian activation \cite{BagIcasspPBN} (similar to softplus) :
$\lambda_{TG}(\alpha)=\alpha + \frac{{\cal N}(\alpha)}{\Phi(\alpha)}$, where
${\cal N}\left(x\right) \defined \frac{e^{-x^2/2}}{\sqrt{2\pi}}$ and $\Phi\left( x\right)  \defined \int_{-\infty}^x {\cal N}\left(x\right).$ 
The TG activation is the theoretical activation function for reconstructing
data from a linear transformation applied to positive-valued data under
the truncated Gaussian prior distribution \cite{BagIcasspPBN}, however in behavior, it is 
similar to softplus $\lambda_{SP}(\alpha)=\frac{1}{1+e^{-x}}$.

\subsection{Classification Results}
The networks were first initialized with random weights and 
trained as a standard discriminative deep neural network (DNN)
with dropout regularization of 0.2, 0.1,  and 0.1 applied to the output of
the first through third layers (dimensions 360, and 100, 36, respectively)
and L-2 regularization.  
No data augmentation (such as random shifting) was used.
Classification accuracy for the DNN is given in Table \ref{tab1}.
\begin{table}[htb]
\begin{center}
 \begin{tabular}{|l|l|l|l|l|l|l|}
\hline
 & \multicolumn{2}{|c|}{"three-tree"} &  \multicolumn{2}{|c|}{"no-go"}  & \multicolumn{2}{|c|}{"bird-bed"} \\
 \hline
	 & train & test & train & test & train & test  \\
 \hline
	 DNN & 1.00 & 0.870 & 1.00 & 0.874 & 1.00 & 0.960 \\
 \hline
	 PBN & .991 & 0.881 & 0.992 & 0.810 & 0.978 & 0.946 \\
 \hline
\end{tabular}
\end{center}
	\caption{Classification accuracy for the thre class pairs.}
	\label{tab1}
\end{table}

Next, the DNNs trained as described above were used as the initial network
for the PBN, which was trained by maximizing the mean likelihood function (\ref{cr1a})
with output prior distribution parameter  $C=2000$.
No data augmentation was used, and no regularization was used.

The classification results for the PBN on the three class pairs are shown
in Table \ref{tab1} where they can be compared with the DNN. Note that
the PBN has only slightly lower accuracy than the DNN,
and in fact has higher accuracy on one class pair.
It should also be kept in mind that the PBN was trained without
regularization of any kind, wheras the DNN used both L2 and dropout
regularization.

\subsection{Reconstruction Results}
It has been established that the PBN has lost little in terms of
classification performance. It will now be determined
what has been gained in terms of generative power.
The first thing that comes to mind is the reconstruction of visible data
from the hidden variables.  Using the method of Section \ref{reconsec},
we reconstructed data from the DNN's hidden variables,
starting with 1 layer, then 2 layers, etc.
Results are shown in Figure \ref{dnnrecon}.
\begin{figure}[h!]
  \begin{center}
    \includegraphics[width=3.5in]{dnnrecon1.eps}
  \caption{Samples of spoken word commands 
	``three" and ``tree" reconstructed from standard DNN hidden variables. From top: original spectrograms,
then the same reconsructed from first through third layer.
Reconstruction from layers 4 and 5 was not possible due to 
sampling efficieny of zero.}
  \label{dnnrecon}
  \end{center}
\end{figure}
After the first layer, some resemblance can be seen, 
but after that, the images are unrecognizeable.
This is how the network sees the data through the hidden variables.
The noisy images, when used as input data will produce exactly
the same hidden variables at the given layer as the
original input sample.

The reconstruction experiment was repeated for the trained PBN.
Results are shown in Figure \ref{pbnrecon}. 
\begin{figure}[h!]
  \begin{center}
    \includegraphics[width=3.5in]{pbnrecon1.eps}
  \caption{Samples of speech commands ``three, three" reconstructed using PBN. From top: original spctrograms, then the same reconsructed 
from output of first through fifth layer.
	  Hidden variable dimensions are 360, 100, 32, 16, and 2.}
  \label{pbnrecon}
  \end{center}
\end{figure}
Reconstructing from the output of the layers produces reconstructions with excellent
quality, but gradually decreasing sharpness.  
Reconstruction from the fifth layer (a feature bottleneck of dimension 2)
produces either a ``standard" ``three" or a ``standard" ``tree".
Note that this network was not trained for lower reconstruction error, but instead to maximize (\ref{cr1a}).
The reconstruction power of the network comes as a side-effect and can be tapped 
into anywhere in the network.  A standard auto-encoder, in contrast,
is trained to reconstruct for a fixed network length.

\subsection{Classifying between class pairs}
A second exercise in ``generative capability" is 
the classification between class pairs using models trained separately
on just one pair. This exercise is obviously not possible
using the discriminative classifiers trained on the
class pairs, but must be trained on all six classes.
For the PBN, discrimination between members of the class pairs
occurs in the output layer and does not require
computing the LF because these are independent terms in
(\ref{cr1a}). To compute the LF for just one member class,
it is necessary to set the parameter $\balpha$ of the
output distribution to the respective one-hot encoding.  The results of the 6-class
PBN experiment are shown in Figure \ref{sixclass}.
The probability output of the PBN classifier
is shown on the left side. Notice that there
are a significant number of errors between class pairs.
Total classification error was 21 percent.

A discriminative DNN was trained on the 6-classes
using the same network structure as the two-class networks, but with increased
neuron counts of 48, 150, 48, 24, and 6 using
dropout and L2 regularization.
The classification performance of the DNN was 12.9 percent,
significantly lower than the multiple PBN classifier.
It is easy to explain why the PBN performed almost as well
in the class-pairs, but much worse in the 6-class experiment.
This is because the class-pair PBN is a single model,
whereas for the 6-class problem, three separate generative
models are required. This introduces errors resulting from imbalances in the models.
Here is an opportunity, however, to improve the 
result by mixing the two classifier outputs.
On the right side of Figure \ref{sixclass}, we show the
classification error percentage as a function of the mixing constant,
showing the transition from PBN only (left) to DNN
only (right).  At a certain value, there is a dip in error,
providing the optimal mixing value.
\begin{figure}[h!]
  \begin{center}
    \includegraphics[width=3.5in]{sixclass.eps}
  \caption{Left: PBN output probabilities for the six classes. Right:
	  Classification error in percent as a function of mixing
	  constant for a mixture of PBN with DNN.}
  \label{sixclass}
  \end{center}
\end{figure}
As has been shown \cite{BagUMS}, a hybrid generative model
can be created at this ``optimal" point.

\subsection{Random Synthesis}
As a final demonstration of generative power, we synthesized entirely random
events by starting with random data equal in dimension to the PBN
output layer, in this case dimension-2.
Data was synthesized at the point prior to the output
activation function using Gaussian random variables.
Results are shown in Figure \ref{syn45} for the class
pair ``three" and ``tree".
\begin{figure}[h!]
  \begin{center}
    \includegraphics[width=3.5in]{syn45.eps}
  \caption{Top: ten training samples randomly selected from
	   ``three" and ``tree" spoken word commands. Bottom: randomly synthesized
	   data from trained PBN. There is no relationship to the
	   selected training samples on top.}
  \label{syn45}
  \end{center}
\end{figure}
The synthetic samples appear realistic and are diverse,
showing variations in time shift, dilation, and other qualities.
This means that the PBN has indeed learned much about the
data generation process.

\subsection{Implementation and Applications }
The PBN was implemented in Python using Theano 
symbolic expression compiler \cite{Theano}.
The primary computational challenge is the solution
of a symmetric linear system with dimension $M\times M$,
where $M$ is the total output dimension of a layer.
This must be solved for each iteration in the solution
of (\ref{hsola}).  This was parallelized on the GPU, one processor
for sample in a mini-batch.  The computational time for an epoch was 1.1 seconds.
This was only about an order of magnitude slower than training the DNN.
All results were obtained using PBN Toolkit \footnote{http://class-specific.com/pbntk. A copy
of the data is also available at this link}.

\section{Conclusions and Future Work}
In this paper, a projected belief network (PBN), which is a purely
generative layered network,  was trained as a
generative-discriminative classifier.  This was achieved using a label-dependent prior for the output features.
Since the PBN is based on a feed-forward neural network (FF-NN),  it can share
an embodiment with a discriminative deep neural network (DNN). Using
a single parameter, the network can be trained either as a generative 
PBN, or as a discriminative DNN, or any point in between. 
When reconstructing visible data from the hidden variables, it was shown that the
DNN had very poor ability to reconstruct, even from initial layers,
whereas when training jointly with the PBN, the reconstruction greatly improved.
The PBN classifier had comparable classification performance
to the discriminative DNN despite using no regularization, yet provided generative power
from three standpoints: visible data reconstruction from hidden variables,
random data synthesis, and classification of out-of set samples.

The results in this paper open up several questions for future work. 
For example, the PBN appears not to respond well to dropout or L2 regularization,
but the unregularized PBN classifier performed on par with the regularized DNN.
Is there a way to regularize the PBN to achieve further improvements?
How can the strucure of the PBN be improved, for example with longer or shorter
networks?  Can the PBN be used in adversarial networks? Are there other, better
prior distributions to use?  Are there better methods to initialize the PBN?
How does classification through reconstruction error compare to LF classification?

\bibliographystyle{ieeetr}
\bibliography{ppt}
\end{document}

\subsection{Background and Motivation }
Discriminative neural networks have dominated machine learning for decades.
The performance of generative networks lags behind 
because they need to model the generative process underlying the data, a much harder
task than discrimination \cite{Vapnik99}. Yet, interest in generative models persists
because a model of the underlying process is useful, as exemplified by variational
autoencoders (VAE) \cite{pmlr-v32-rezende14}, and generative adversarial network \cite{GoodfellowGAN2014}
(GAN) which have sparked considerable interest.
While the generative task is harder, given time and effort, 
generative models can perform as well as classifiers as
their discriminative counterparts. 
For example, when Hinton's deep belief network (DBN) 
was published, the DBN worked better than comparable fully-connected 
(non-convolutional) feed-forward networks \cite{HintonDeep06}.
While training algorithms have been developed for VAE and DBN,
the likelihood functions (LF) are not available in closed-form, so need to be approximated,
using stochastic variational methods in the case of VAE \cite{pmlr-v32-rezende14},
or Monte Carlo approximations in the case of DBN \cite{SalakhutdinovDBN}.
%
%
The projected belief network (PBN) is 
a new type of generative network with tractable 
LF that generates data layer-wise from hidden variables similar to a 
 deep latent Gaussian model (DLGM).  But, in contrast to other
generative models, the PBN  is related to a feed-forward neural 
network (FF-NN)  by a duality relationship \cite{BagPBN}.  
The dual FF-NN, which is here called dual analysis network (DAN), 
exactly recovers the hidden variables
of the PBNs data generation process.  
%
%
With tractable LF, the PBN has the potential to enable a new
class of generative models and algorithms.

\subsection{Main Idea}
The projected belief network (PBN) was previously introduced as a dual counterpart
to a feed-forward neural network (FF-NN) \cite{BagPBN}.
The PBN is derived from a FF-NN by asking the following question: {\it knowing
the FF-NN and the distribution of the output variables (features) of the FF-NN,
what is the  maximum entropy (MaxEnt) distribution of the visible 
data consistent with the given features distribution?}
The PBN is the generative network that implements this MaxEnt distribution \cite{BagPBN}.
Not surprisingly, the PBN uses the same network weights as the FF-NN
from which it is derived, and employs a special 
``activation" function that gives it its unique properties.
A deterministic version of the PBN is created if instead of
generating random data in each layer, the conditional
mean is propagated.  The deterministic PBN is the complementary
network to the DAN and combined with the DAN forms a new
type of auto-encoder.

\subsection{Paper Contributions}
The PBN has been previously introduced \cite{BagPBN}. Novel contributions of this paper include
(a) experimental results comparing PBN with 
other models as a function of data dimension,
(b) the detailed description of a multi-layer PBN,
(c) the treatment of the issue of sampling efficiency,
(d) the conceptual comparison of PBN with the VAE,
and (e) the description of a deterministic PBN
and its application as an auto-encoder,
and experiments showing significant improvements
over a conventional auto-encoder of the same structure.

\section{Projected Belief Networks (PBN)}
\subsection{PBN Exact Form}
Figure \ref{pbn_multi} illustrates a two-layer PBN in its exact, asymptotic,
and deterministic forms.  It can be easily extended to more layers.
\begin{figure}[h!]
  \begin{center}
    \includegraphics[width=3.5in]{pbn_multi_b.eps}
  \caption{A 2-layer PBN in three forms,
exact, asymptotic, and deterministic, and the
corresponding dual analysis network (DAN).}
  \label{pbn_multi}
  \end{center}
\end{figure}
Near the bottom of the figure is the dual analysis network (DAN), a conventional
feed-forward network employing an activation function
$\lambda_n(\;)$ in layer $n$.
Optionally, an energy statistic (ES), denoted by $e=t(\bfx)$ is extracted from the input
of each layer.
The figure illustrates both data generation by different forms of the
PBN (left to right) and feature extraction by the DAN (right to left).
Data generation originates by a feature generating distribution
$g(\bfz_2)$, then continues layer by layer.  In layer $n$ of the exact form of the PBN, (top), the
activation function and bias (if used) are inverted, and the
feature $\bfz_n$ is presented to the ``UMS" block in which a 
sample $\bfx$ is drawn randomly from the set ${\cal M}_n(\bfz_n,e_n)$ defined by
\beq
   {\cal M}_n(\bfz_n,e_n) = \{ \bfx : {\bf W}_n^\prime \bfx = \bfz_n, \; t_n(\bfx)=e_n, \;\;\; \bfx \in {\cal X}_n \},
   \label{manifze}
\eeq
where ${\cal X}_n$ is the input range of layer $n$ and $e_n=t_n(\bfx)$ is the optional ES.
The sample $\bfx$ must be drawn with uniform distribution,
so that no member of ${\cal M}_n(\bfz_n,e_n)$ is more likely to be drawn than any other.
The sampling procedure is therefore called uniform manifold sampling (UMS) \cite{BagUMS}.

By the definition of UMS, the DAN will exactly recover the variables $\bfz_2$, $\bfz_1$. 
When the PDF of $\bfz_2$ is known, denoted
by $g(\bfz_2)$, then the PBN generates samples the PDF:
\beq
p_p(\bfx_1; T, g) = \frac{1}{\epsilon} \; \frac{p(\bfx_1 ; H_{0,1})}{p(\bfz_1 ; H_{0,1})} \;  |{\bf J}_{\bfz_1 \bfx_2}| \; \frac{p(\bfx_2 ; H_{0,2})}{p(\bfz_2 ; H_{0,2})} 
\; g(\bfz_2),
\label{cr1a}
\eeq
where $\bfx_n$ is the input data to layer $n$ ($\bfx_1$ is the visible data), 
$T$ represents the DAN, $|{\bf J}_{\bfz_1 \bfx_2}|$ is the determinant of the
Jacobian of the 1:1 mapping from $\bfz_1$ to $\bfx_2$,
and $\epsilon$ is the sampling efficiency, to be explained below.

Notice the absence of integral signs in (\ref{cr1a}) - the distribution
does not require integrating out the hidden variables, as is necessary
in other layered generative models.  This is due to the fact that the
hidden variables of the DAN are deterministically derived from the
visible data, not jointly distributed.   Note also that in (\ref{cr1a})  
there appears a set of reference distributions, one for each layer.
The distribution $p(\bfx_n;H_{0,n})$ is the maximum entropy (MaxEnt) reference distribution
for layer $n$ and $p(\bfz_n;H_{0,n})$ is the corresponding feature distribution\footnote{$p(\bfz_n;H_{0,n})$ is the theoretical PDF
of the layer output when the layer input is distribued according to $p(\bfx_n;H_{0,n})$.}.
This reference distribution depends on ${\cal X}_n$, the data range of layer $n$ input,
which in turn depends on the activation function used in the previous layer - note
that the input (visible data) is assumed to have been created using $\lambda_1(\;)$.
We consider three data ranges: $\mathbb{R}^N$, $\mathbb{P}^N$,
and $\mathbb{U}^N$, where $N$ represents input data dimension of a generic layer,
$\mathbb{R}^N$ is the unlimited case, $\mathbb{P}^N$ is the positive quadrant ($0\leq x_i$),
and $\mathbb{U}^N$ is the unit hypercube ($0\leq x_i \leq 1$).
%
The MaxEnt reference distribution for each data range ${\cal X}$ is given in Table \ref{tab1v}.
The primary computational challenge in computing (\ref{cr1a})
is calculating the denominator terms $p(\bfz_n;H_{0,n})$.  More is provided in the 
references \cite{BagPDFProj,BagNutKay2000,Bag_info,BagUMS,BagPBN,BagEusipcoRBM}.
\begin{table}
\begin{center}
 \begin{tabular}{|l|l|l|l|l|}
\hline
${\cal X}$ &  $p(x;\alpha)$ & $\lambda(\alpha)$  & $t(\bfx)$ & $p(\bfx;H_0)$\\
 \hline
$\mathbb{R}^N$   &  ${e^{-(x-\alpha)^2/(2\sigma^2)} \over \sqrt{2 \pi \sigma^2} }$ (Gauss.) & $\alpha$  & $\sum_i x_i^2$  
& $\frac{e^{-t^2(\bfx)/2}}{(2\pi)^{-N/2}}$\\
 \hline
$\mathbb{P}^N$  &  $\alpha e^{-\alpha x}$  {\hspace{.37in}} (Expon.) & $1/\alpha$    & $\sum_i x_i$  & $e^{-t(\bfx)}$ \\
 \hline
$\mathbb{U}^N$   &   $\left(\frac{\alpha}{e^{\alpha} - 1}\right)  \; e^{\alpha x}$   {\hspace{.12in}} (TED) & $\frac{e^{\alpha}}{e^{\alpha} - 1}-\frac{1}{\alpha}$    & none & 1\\
 \hline
\end{tabular}
\end{center}
\caption{Generating distributions $p(x;\alpha)$, expected value of generating distributions $\lambda(\alpha)$,
energy statistics (ES) $t(\bfx)$,  and reference hypotheses $p(\bfx;H_0)$
for for data ranges $\mathbb{R}^N$, $\mathbb{P}^N$, and $\mathbb{U}^N$.
This table concerns a single layer and ${\bf x}$ is assumed to be the
visible data for the given layer layer with dimension $N$ and range ${\bf x}\in {\cal X}$.
}
\label{tab1v}
\end{table}


Depending on the data range (see Table \ref{tab1v}) an ES might need to be extracted from each layer input. We describe the ES for completeness,
but no ES is needed for $\mathbb{U}^N$, and for $\mathbb{P}^N$, the 
ES can be incorporated into matrix ${\bf W}_n$, eliminating the need for an explicit ES.  For more about the ES, please consult the references \cite{Bag_info,BagUMS}.

Optionally, a bias and activation function can be appended to the DAN (bottom of Figure \ref{pbn_multi}),
producing feature $\bfx_3$. In this case, the data generation process begins with the generating distribution
$g(\bfx_3)$, and the activation function and bias must be inverted.  Also, 
equation (\ref{cr1a}) must be modified by replacing $g(\bfz_2)$ with $|{\bf J}_{\bfz_2 \bfx_3}| \;g(\bfx_3).$

%

\subsection{PBN Asymptotic Form}
It has been shown that the UMS sampling process can be closely approximated by a
network layer resembling a sigmoid belief network \cite{BagUMS}.
To arrive at the asymptotic PBN (see Figure \ref{pbn_multi}), 
the UMS blocks are replaced by a nonlinear function $\bfh_n = \gamma_n^{-1}(\bfz_n)$,
 matrix multiplication $\balpha_n = {\bf W}_n \bfh_n$, then generation
from distributions $p_n(x; \alpha)$, which
are given in Table \ref{tab1v} as a function of ${\cal X}_n$.  The expected value of these distributions
(given $\alpha$) is denoted by $\lambda_n(\alpha)$, which
corresponds to the activation functions used in the DAN
at the output of layer $n-1$.  
Interestingly, for $\mathbb{U}^N$, $\lambda_n(\alpha)$
 is the mean of the truncated exponential distribution (TED),
which is similar to the sigmoid function \cite{BagUMS}.
Central to the theoretical analysis of a PBN layer
is the function $\gamma_n(\bfh_n)  =  {\bf W}_n^\prime \lambda( {\bf W}_n \bfh_n).$
To compute a layer of a PBN, this function needs to be inverted:
$\bfh_n=\gamma_n^{-1}(\bfz_n),$ which requires  
an iterative algorithm, but
might have no solution (See Section
\ref{sampeff}).

\subsection{The PBN for $\mathbb{R}^{N}$ and Relationship to VAE}
The VAE is currently a well-studied generative model \cite{Goodfellow2016,pmlr-v32-rezende14}.
The ``variational" aspect of VAE has to do with approximating and training the LF,
but the VAE is essentially an implementation of DLGM \cite{pmlr-v32-rezende14}.
Thus, both PBN and VAE are layered generative models. The main difference
is that the PBN is based on an explicit feed-forward analysis network (the DAN),
so the latent variables can be  deterministically computed from the
visible data. So, once a visible data
sample has been generated by the PBN, all the hidden variables can then be exactly
recovered by a single pass of the DAN.
The VAE on the other hand is a stochastic layered generative model,
so the latent variables of the VAE are jointly distributed
with the visible data. For this reason the LF of the VAE is only available 
as an integral over the hidden variables.
But, this distinction is moot because when looking at the asymptotic form of the PBN,
an approximation that is very good as has been demonstrated \cite{BagUMS},
we see that the PBN {\it behaves} like a traditional layered stochastic generative model.

A network layer of a DLGM is composed of an arbitrary
non-linear function followed by additive correlated noise \cite{pmlr-v32-rezende14}.
A network layer of an asymptotic PBN, on the other hand,  is composed of 
a non-linear function $\gamma_n^{-1}(\bfz)$, followed by multiplication by matrix 
${\bf W}_n$, then the generating distributions are applied
to produce the output variables.  Function $\gamma_n^{-1}(\bfz)$ and the generating distributions  
depend on the range of the layer output variable and are given in Table
\ref{tab1v}.  When $\bfx \in \mathbb{R}^{N}$, the generating distribution is
Gaussian, and is implemented by adding independent Gaussian noise\footnote{This can be
easily extended to correlated noise by introducing a matrix multiplification 
between the layers.}.  This produces a type of DLGM.
But, the Gaussian noise in an asymptotic PBN must be added
after a linear transformation, whereas for DLGM it is added after an arbitrary transformation.
It is not clear what this distinction means to the ultimate
PDF estimation capability, and can only be discovered by future experiments. 
Note also that for the DLGM, the activation function is taken to be
part of the ``arbitrary non-linear function" , whereas in the 
PBN, the activation function $\lambda_n(\;)$ is defined for the
dual DAN, which determines the function  $\gamma_n^{-1}(\bfz)$ used in the PBN.
In holding to the MaxEnt principle, 
for a given data range ${\cal X}$, the activation function 
$\lambda()$ is fixed, and therefore $\gamma_n^{-1}(\bfz)$ is fixed.
But, if one is willing to give up this MaxEnt distinction, there is 
flexibility in choosing $\lambda()$  so long
as it is invertible (for example use {\it softplus}, not {\it relu}).

In summary, both DLGM and PBN are layered
generative networks and it is not clear from the
above comparison which structure is better or more general.
It is clear, however, that the PBN under special conditions
(i.e. for ${\cal X}=\mathbb{R}^N$) approximates a type of DLGM and has a closed-form LF
 which is especially efficient to compute for this case (see \cite{Bag_info} Section IV.C, page 2821).
Future work is planned to compare DLGM and PBN in practice.

\subsection{PBN Deterministic Form}
The deterministic form of the PBN is obtained from the
asymptotic form by replacing $p_n(x;\alpha)$ by their expected values $\lambda_n(\alpha)$.
Interestingly, $\lambda_n(\alpha)$ cancels  $\lambda^{-1}_n(\alpha)$,
leaving $\gamma^{-1}_n(\;)$ as the only non-linearities, except at the visible layer.
This resulting PBN is a deterministic dual to the DAN, which exactly recovers the hidden values.
An arbitrary activation function $\lambda_n(\alpha)$ can be used as long
as $\gamma_n(\bfh_n)$ is defined using the same function. 
Note that $\lambda_n(\alpha)$ must be invertible, so
activations functions like {\it softplus} can be used, but not {\it relu}.

\subsection{Sampling Efficiency}
\label{sampeff}
The sampling efficiency $\epsilon$ is the fraction of times that
the PBN successfully creates a sample of visible data and
depends on the feature generating distribution
$g(\bfz)$ and whether exact (UMS) or deterministic generation is used.
A sampling failure occurs in a UMS block if the set
${\cal M}_m(\bfz_n,e_n)$ has no members, or in the asymptotic or
deterministic PBN if $\gamma^{-1}_{n}({\bf z}_n)$ has no solution.
When sampling fails, it is necessary to re-start
the process by drawing another feature value.
Sampling efficiency, either for UMS or
for deterministic PBN, can be driven towards 1.0 though
training, as will be demonstrated below.

\subsection{PBN Initialization and Training}
\label{jft}
In order to initialze the PBN so it has high sampling efficiency,
the weight matrices should be initialized by principal component analysis (PCA)
of the input data prior to the activation function \footnote{When data is already 
constrained to the range $[0,\; 1]$, as it is
in the MNIST corpus, it is useful to ``gaussianify" the data, mapping to $\mathbb{R}^N$ prior to PCA analysis (See Section \ref{ddesc}).}.
Scaling and bias are then used to provide good ``activation" of $\lambda_n(\;)$.
%
%
In this paper, two types of PBN training are used - deterministic auto-encoder training
and maximum likelihood (ML) training.
In auto-encoder training, the 
DAN is combined with the deterministic PBN to form an auto-encoder (a clockwise circular path at the 
bottom of Figure \ref{pbn_multi}).  Training is accomplished using back-propagation
to minimize total square reconstruction error.
Note that the parameters appear in both PBN and DAN, so the derivative has two terms.
It is critical to have high sampling efficiency for 
auto-encoder training.  In the experiments, $\epsilon$ 
approaches very nearly 1.0 after the first training epoch, 
even for testing data.


In ML training, the log of equation (\ref{cr1a}) is trained for highest average value
by gradient ascent.
%
%
%
We used a special ``uniform assumption" training in which 
the optional activation function (bottom of Figure \ref{pbn_multi}) is applied
to compress the data to the range [0,1],  
and the feature distribution $g(\bfx_3)$ is ignored.
Ignoring the feature distribution is tantamount
to assuming that $g(\bfx_3)=1$, the uniform distribution.
Interestingly, by training this way, a network is produced that,
in fact, produces feature data $\bfx_3$ that is independent uniformly distributed - the 
simplifying assumption becomes fulfilled.

\section{Classification Experiments}
We now compare PBN with a Gaussian mixture model (GMM) in a simple 
classification task.

\subsection{Reduced MNIST Data Description}
\label{ddesc}
For the following experiments, just three characters
``3", ``8", and ``9", of the MNIST handwritten data corpus were used.
Four pixel down-sampling rates were chosen: 1:1, 2:1, 3:1, and 4:1,
resulting in 
data dimensions of 784, 196, 100, and 49.
Since MNIST pixel data is coarsely quantized in the range [0,1],
a dither was applied to the pixel values\footnote{For pixel values
above 0.5, a small exponential-distributed random value was subtracted,
but for pixel values below 0.5, a similar  random value was 
added.}.  To create data in $\mathbb{R}^N$, the inverse sigmoid function was
then applied in order to create ``gaussianified" data with most pixel values in the range -10 and 10.  

\subsection{The 1-layer PBN}
We revisit the 1-layer PBN, which was previously introduced
\cite{BagPBN}. The results of 1-layer PBN experiments 
are relevant to determine if the PBN should be exended to a second layer.
In a multi-layer PBN, a given layer acts as a PDF
model for the features of the up-stream layer. So,
it seems that there is no advantage to adding a layer to a PBN
if a GMM works better than the added layer.
The idea, then is to test a 1-layer PBN against
a GMM as a function of dimension.
This experiment is data-set dependent, so the results
here apply only to MNIST.
%
As a performance benchmark, the GMM was 
applied to the ``gaussianified" data in $\mathbb{R}^N$,
using both diagonal (GMM-D), and full (GMM-F) covariance matrices
\footnote{To avoid singularities, the diagonal elements of the covariance
matrices were multiplied by the factor $(1+\delta)$, where 
$\delta=$ 0.3, 0.3, 0.5, and 0.6 for dimensions 49, 100, 192, and 784, respectively.}.
A separate 1-layer PBN was initialized using PCA,
then trained for each data class 
to maximize the mean log-likelihood using gradient ascent
with ``ADAM" optimization and  L2 regularization using ``uniform assumption" 
training (Section \ref{jft}).  After training, the final activation function was removed,
then $g(\bfz_1)$ was modeled as a GMM.
%
For $N=49, 100, 196, 784$, the number of hidden units
(columns of matrix ${\bf W}$) were 12, 16, 30, and 34, respectively.

Results of the experiment are shown in Figure \ref{pbn_N}.
The PCA-initialized PBN, with no further training
are reported as ``PBN-P", and with training as ``PBN-G".
When comparing ``PBN-P" with ``PBN-G", we can conclude
that ML training greatly improves a PBN.  This means that 
the PDF model offered by a 1-layer PBN is more than 
a just a re-packaged type of Gaussian model or PCA.
The next observation is that the PBN performs better than GMM-F above $N=100$.
%
%
\begin{figure}[h]
  \begin{center}
    \includegraphics[width=3.4in,height=2.4in]{pbn_N.eps}
  \caption{Model comparison as a function of data dimension.} 
  \label{pbn_N}
  \end{center}
\end{figure}
%
%
Both  GMM-F and PBN can model pixel correlation, GMM-F explicitly
using the covariance matrices, and PBN implicitly 
by decorrelating the features, as was noted at the end of Section \ref{jft}.
But, PBN requires $MN$ parameters, versus  the $MN^2$ parameters required for the GMM.
This may explain the advantage of PBN above $N=100$.  
The average sampling efficiency for PBN-G 
was 0.72, 0.85, 0.77, and 0.52 for $N=49,100,192,784$,
respectively.  The worst case change in per-pixel log-likelihood,
is 0.007, so sampling efficiency in Figure \ref{pbn_N} can be essentially
ignored.

\subsection{Multi-layer PBN}
The 1-layer PBNs for $N=196$ and $784$
were extended to a second layer with $16$ and $18$ hidden units, respectively.
The 2-layer PBNs were then trained with an assumption of uniform distribution for $g(\bfx_3)$, then the final activation function was removed and
GMM was used to model the final feature PDF $g(\bfz_2)$.  
Sampling efficiencies were 0.55 and 0.70, respectively,
also negligible.  Performance is shown in Figure \ref{pbn_N} as ``PBG-2-G"
and shows worse performance with respect to 1-layer PBN-G.
This could have been predicted based on Figure
\ref{pbn_N} because the feature dimension is much less than $100$.
Extending the PBN to a second layer would only be effective if the
first layer feature dimension is much larger. 
%

\section{Auto-Encoder (A-E) Experiments}
In the next experiment, a multi-layer
deterministic PBN together with the DAN
are used as an A-E and compared with a standard A-E network 
of the same structure. 
The full $28\times 28$ ($N=784$) data was used.
Separate A-Es were trained on each data 
class to minimize total square error by back-propagation.
ADAM optimization and L2 regularization was used for both network types.
TED (T), sigmoid (S) and softplus (P) activation functions were tried.
The average squared error was measured for testing and training data
and is listed in Table \ref{tab1a}. Although the 
conventional A-E attained a lower squared error
on the training data, it fared much worse on the test data. In contrast,
the PBN had similar squared error on both sets, significantly
out-performing the standard A-E - which
can probably be attributed to (a) that fact that the 
PBN uses the same weights for reconstruction and analysis, and thereby
implements the same task with half the parameters, and (b)
the reconstruction (PBN) is the perfect complement to the 
analysis network (DAN).
Using L2-regularization for conventional A-E did not change this.
The A-E performance for TED and sigmoid was similar, but training took longer for TED.
Sampling efficiency for PBN was 100 percent (no samples that failed reconstruction)
for training, and about 99.9\% (typically 1 sample or less failed) on the
test data.  
\begin{table}
\begin{center}
 \begin{tabular}{|l|l|l|l|l|l|l|}
\hline
Nodes & Act & Type & E-Train & E-Test & Class  \\
\hline
\hline
32-12 & T & A-E  & 7.40 & 10.39 & 1.94\% \\
\hline
32-12  & S & A-E  & 6.73 & 10.79 & 2.97\% \\
\hline
32-12 & T & {\bf PBN}  & 8.63 & {\bf 9.04} & {\bf 1.27\%}  \\
\hline
\hline
36-16 & T & A-E  & 5.84 & 8.21 & 2.57\% \\
\hline
36-16  & S & A-E  & 5.26 & 8.26 & 2.51\% \\
\hline
36-16 & T & {\bf PBN}  & 6.96 & {\bf 7.40} & {\bf 1.70\%}  \\
\hline
\hline
32-16-9 & P & A-E  & 8.27 & 15.3 & 4.4\%  \\
\hline
32-16-9 & P & {\bf PBN}  & 9.95 & {\bf 11.25} & {\bf 0.90\%}  \\
\hline
\end{tabular}
\end{center}
\caption{Total square error for auto-encoder task. Activation functions
(Act) are TED (T), sigmoid (S) and softplus (P)}.
\label{tab1a}
\end{table}
The good generalization of the PBN A-E suggests
using it as a classifier based on minimum reconstruction
error, which we tried.  The results are shown in Table \ref{tab1a} 
in column ``Class".  PBN performed significantly better than A-E,
attaining a very respectable 0.9\%, which handily out-performs 
the standard PBNs in Figure \ref{pbn_N} (denoted by ``PBN A-E").

The deterministic PBN is also useful to generate entirely synthetic data,
In Figure \ref{aenc_syn_pbn}, examples were generated by training a 
GMM on the features (i.e. output of the DAN), then 
passing synthetic features through the PBN. 
The configuration ``32-16-9" with softplus activation was used.  The synthetic samples are sorted in order of decreasing likelihood
(starting from top left), demonstrating the a benefit of 
a tractable likelihood function.
The quality of these samples suggests using the deterministic PBN
in a generative adversarial network (GAN)  - but differing from
a standard GAN in the posession of a tractable LF.
\begin{figure}[h]
  \begin{center}
    \includegraphics[width=3.5in]{aenc_syn_pbn.eps}
  \caption{Data synthesized from determinisic PBN and sorted in order of decreasing
likelihood value.}
  \label{aenc_syn_pbn}
  \end{center}
\end{figure}

\ifdohybrid
\section{Hybrid PBN classifier}
The goal in this experiment is to combine the
results of the last 2 sections by forming a hybrid PBN classifier 
from the 1-layer PBN classifier and the PBN auto-encoder.

For the PBN portion of the hybrid, we used an annealed kernel mixture.
The idea of a PBN kernel mixture is that the features extracted by
a PBN trained on one class might have useful information
regarding another class - especially for poorly formed 
handwritten characters.
The class-specific feature 
mixture (CSFM) \cite{BagIWCCSP,BagAESModelMix,BagUMS}  
approximates the PDF of one class using a mixture of 
all the PBNs, increasing the information
available without increasing the feature dimension.
Annealing improves the linear mixing of the kernels
\cite{BagAESModelMix,BagUMS}.  
We form an annealed kernel mixture of the PBN PDFs (\ref{cr1a}) 
as follows:
\beq
f_m(\bfx; a) = \left(
\sum_{l=1}^c w_{l,m} \; p_p(\bfx; T_l,\smallmath{\hat{p}(\bfz_l|H_m)})^{1/a}
\right)^a,
\label{csfma}
\eeq
where $c$ is the number of classes ($c=3$ here), 
$\bfz_l = T_l(\bfx)$ represents the DAN trained on class $l$,
$\hat{p}(\bfz_l|H_m)$ is the feature PDF estimate for feature $\bfz_l$ and data class $m$,
and $a$ is a heurisic annealing parameter.
The weights are estimated using training data using,
$$\hat{w}_{l,m} = \frac{ \sum_i \; p_p(\bfx_i; T_l,\smallmath{\hat{p}(\bfz_l|H_m)})^{1/a}}
{\sum_i \; \sum_k \; p_p(\bfx_i; T_k,\smallmath{\hat{p}(\bfz_k|H_m)})^{1/a}}.$$
For the deterministic PBN auto-encoder portion of the hybrid, 
we used $h_m(\bfx;b) = {e^{-\|\bfx-\hat{\bfx}_m\|^2/b} \over
\sum_l e^{-\|\bfx-\hat{\bfx}_l\|^2/b}},$
where $\|\bfx-\hat{\bfx}_l\|^2$ is the square auto-encoding error
using auto-encoder trained on class $l$.
The complete hybrid classifier distribution is given by
$p(\bfx|H_m; a,b)=\frac{f_m(\bfx; a) \; h_m(\bfx;b)}{K_m(a,b)}$, where
$K_m(a,b)$ is the normalization constant 
that can be estimated using Monte Carlo integration (MCI)
with the un-annealed mixture $f_m(\bfx; a=1)$ 
acting as proposal distribution\footnote{ 
As a motivation for PBN, we noted that having a
tractable LF avoids the need for MCI, yet here we are using MCI. 
This is not a contradiction.  As dimension increases, the proposal distribution needs 
to be increasingly well matched to the function to be integrated.
To normalize a high-dimensional distribution outright,
for example using GMM as proposal distribution would fail.
But, MCI is useful to normalize 
high-dimensional distributions with tractable LF 
that have been slightly modified,  where the un-modified
distribution acts as proposal distribution.
} \cite{BagUMS}.

Prior to estimating $K_m(a,b)$, classification error was optimized over $a$ (Figure \ref{comb_b} left), without the term  $h_m(\bfx;b)$.
Then, using the value $a=1500$, optimized over $b$
(Figure \ref{comb_b} right), with a resulting minimum error of 0.77\%,
entered in Figure \ref{pbn_N} (left) as ``PBN-H".
With these values of $a$ and $b$, the normalization constant
$K_m(a,b)$ was estimated using Monte Carlo integration
and the normalized LF entered in Figure \ref{pbn_N} (right).
\begin{figure}[h]
  \begin{center}
    \includegraphics[height=1.9in,width=1.1in]{comb_a.eps}
    \includegraphics[height=1.9in,width=1.1in]{comb_b.eps}
    \includegraphics[height=1.9in,width=1.1in]{comb_b_10.eps}
  \caption{Classification error on reduced MNIST
as a function of  $a$ (left) and  $b$ (center), and on full MNIST
as a function of $b$ (right).}
  \label{comb_b}
  \end{center}
\end{figure}

As a final experiment, the hybrid classifier was tried on the full 10-character MNIST data set
to verify the above results and so that it could be compared with published work.
The classification error as a function of $b$ is shown in Figure \ref{comb_b}, right side, and attains a minumum error of 1.25\%
at the same value of $b$, which is comparable to state of the art
fully-connected (non-convolutional) discriminative classifiers not employing pre-processing,
image distortions or deskewing \cite{MNISTResults}.
\fi

\section{Conclusions}
In this paper, a multi-layer PBN has been described,
in its standard, asymptotic, and deterministic forms.
Experiments comparing a 1-layer PBN with a GMM
on a reduced subset of MNIST show that PBN out-performs GMM 
only above a dimension of about 100, which would
suggest using a 2-layer PBN when the output dimension of the first layer is large.
This paper also described a deterministic multi-layer PBN for the
first time and it has been experimentally found to be superior to a standard auto-encoder
when generalizing to test data both in terms of
reconstruction error and classifier performance.
\bibliographystyle{ieeetr}
\bibliography{ppt}
\end{document}